\documentclass{article}
\usepackage{graphicx}
\usepackage{acl}
\usepackage{times}
\usepackage{graphicx}
\usepackage{booktabs}
\usepackage{caption}
\usepackage{soul}
\usepackage{verbatim}
\usepackage{wrapfig}
\usepackage{siunitx}
\usepackage{longtable}
\usepackage[T1]{fontenc}
\usepackage[utf8]{inputenc}
\usepackage{newtxtext,newtxmath}
\usepackage{caption} 
\usepackage{amsmath,amsfonts,bm}

\def\eqref#1{equation~\ref{#1}}

\def\1{\bm{1}}

\DeclareMathAlphabet{\mathsfit}{\encodingdefault}{\sfdefault}{m}{sl}
\SetMathAlphabet{\mathsfit}{bold}{\encodingdefault}{\sfdefault}{bx}{n}

\newcommand{\misscite}[1]{\textcolor{red}{[CITE]}}
\newcommand{\missingnumber}[1]{\textcolor{red}{[NUMBER]}}

\newcommand{\Sref}[1]{\S\ref{#1}}

\usepackage{enumitem}

\definecolor{codegreen}{rgb}{0,0.6,0}
\definecolor{codegray}{rgb}{0.5,0.5,0.5}
\definecolor{codepurple}{rgb}{0.58,0,0.82}
\definecolor{backcolour}{rgb}{0.95,0.95,0.92}

\newcommand{\applygradient}[1]{%
  \pgfmathparse{#1}%
  \let\val=\pgfmathresult 
  \pgfmathparse{int(\val)} 
  \let\intval=\pgfmathresult 
  \ifnum\intval<50
    \pgfmathsetmacro{\r}{255}
    \pgfmathsetmacro{\g}{128 * \val / 50}
  \else
    \pgfmathsetmacro{\r}{255 * (100 - \val) / 50}
    \pgfmathsetmacro{\g}{255}
  \fi
  \pgfmathtruncatemacro{\rint}{\r}%
  \pgfmathtruncatemacro{\gint}{\g}%
  \xdef\colorname{\noexpand\cellcolor[RGB]{\rint,\gint,0}}
  \colorname
  \val
}

\usepackage{hyperref} 
\usepackage{url}
\usepackage[dvipsnames, table]{xcolor}
\usepackage[most]{tcolorbox} 
\usepackage{listings}
\usepackage{xcolor}
\definecolor{metricBlueBG}{HTML}{FFFFFF}
\definecolor{metricBlueFG}{HTML}{5A80D6}
\definecolor{metricBlueBG2}{HTML}{dde5f6}
\definecolor{metricOrangeFG}{HTML}{d6b05a}
\usepackage[most]{tcolorbox}
\usepackage{multicol}
\renewcommand{\thefootnote}{\fnsymbol{footnote}}
\tcbset{
  colback=metricBlueBG2, 
  colframe=metricBlueFG, 
  boxrule=0.5pt,
  arc=2mm,
  left=2mm,
  right=2mm,
  top=1mm,
  bottom=1mm,
}
\lstdefinelanguage{json}{
  basicstyle=\ttfamily\small,
  breaklines=true,
  breakatwhitespace=true,
  showstringspaces=false,
  frame=none
}

\title{\includegraphics[height=1.1em]{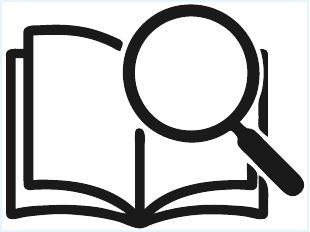} ScholarEval: Research Idea Evaluation \\ Grounded in Literature}

\author{Hanane Nour Moussa$^1$\footnotemark[2],
Patrick Queiroz Da Silva$^1$\footnotemark[2],
Daniel Adu-Ampratwum$^2$, Alyson East$^3$, \\
\textbf{Zitong Lu$^4$, Nikki Puccetti$^5$, Mingyi Xue$^6$, Huan Sun$^1$, Bodhisattwa Prasad Majumder$^7$}, \\
\textbf{Sachin Kumar$^1$} \\
$^1$Department of Computer Science and Engineering, The Ohio State University \\
$^2$Division of Medicinal Chemistry and Pharmacognosy, The Ohio State University \\
$^3$Department of Wildlife, Fisheries, and Conservation Biology, The University of Maine \\
$^4$McGovern Institute for Brain Research, Massachusetts Institute of Technology \\
$^5$Center for Cognitive and Behavioral Brain Imaging, The Ohio State University \\
$^6$Department of Chemistry, University of Wisconsin-Madison \\ $^7$Allen Institute for Artificial Intelligence
}

\usepackage{xspace}
\newtcbox{\metricpill}[1][]{
  on line,
  colback=metricBlueBG,
  colframe=metricBlueFG,
  boxsep=0.2ex,
  left=0.6ex,right=0.6ex,top=0.2ex,bottom=0.2ex,
  arc=0pt,            
  boxrule=0.6pt,      
  nobeforeafter,
  tcbox raise base,   
  enhanced,
  #1                  
}

\newcommand{\metricpillblue}[1]{%
  \metricpill[colback=metricBlueBG,colframe=metricBlueFG]{#1}%
}
\newcommand{\metricpillorange}[1]{%
  \metricpill[colback=metricBlueBG,colframe=metricOrangeFG]{#1}%
}

\newcommand{\hnm}[1]{{\color{black}#1}}

\newcommand{\system}{\textsc{ScholarEval}\xspace}

\newcommand{\benchmark}{\textsc{ScholarIdeas}\xspace}

\begin{document}
\maketitle
\footnotetext[2]{Co-leads. All authors' contributions are detailed in the Contribution section.}
\renewcommand{\thefootnote}{\arabic{footnote}}
\setcounter{footnote}{0}
\begin{abstract}
As AI tools become increasingly common for research ideation, robust evaluation is critical to ensure the validity and usefulness of generated ideas. 
We introduce \system, a retrieval-augmented
evaluation framework that assesses research ideas based on two fundamental criteria: \textit{soundness}---the empirical validity of proposed methods based on existing literature, and \textit{contribution}---the degree of advancement made by the idea across different dimensions relative to prior research. 
To evaluate \system 
, we introduce \benchmark, the first expert-annotated dataset of multi-domain research ideas and reviews, comprised of 117 ideas 
across four disciplines: artificial intelligence, neuroscience, biochemistry, and ecology. Our evaluation shows that \system achieves significantly higher coverage of points mentioned in the human expert annotated rubrics
in \benchmark compared to all baselines. Furthermore, \system is consistently preferred over \hnm{the} strong baseline \hnm{OpenAI Deep Research}, a reasoning and search-enabled agentic system, 
in terms of evaluation actionability, depth, and evidence support. Our large-scale user study also shows that \system significantly outperforms \hnm{OpenAI Deep Research} in literature engagement, idea refinement, and usefulness.
We openly release our code, dataset, and \system tool for the community to use and build on.\footnote{Code and data can be found at \url{https://github.com/skai-research/ScholarEval}}  
\end{abstract}
\section{Introduction}
\label{sec:intro}
\begin{figure}[t]  
    \centering
    \includegraphics[width=1\linewidth]{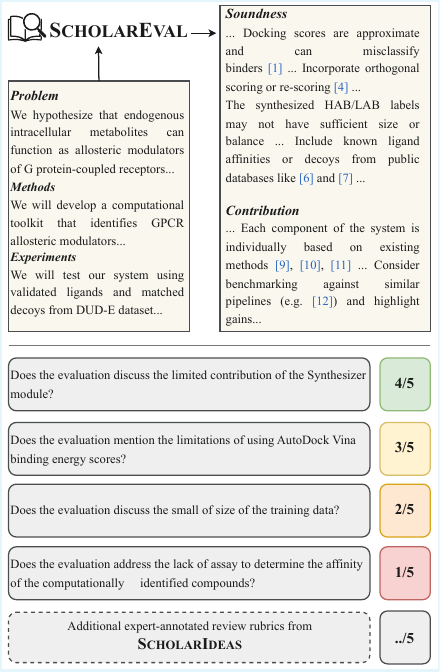}
    \caption{\textbf{Top:} Given a research idea \system generates a literature-grounded evaluation based on soundness and contribution. \textbf{Bottom:} To evaluate \system, we measure the degree of coverage of expert-annotated review rubrics in \benchmark. The final coverage score is the average over all rubrics.}
    \label{fig:highlight}
\end{figure}
Research ideation stands out as one of the most critical and challenging steps in scientific research, where the success of a research project fundamentally hinges on the technical soundness of the underlying idea and its potential to advance the field. To accelerate this stage, multiple works have developed AI-based systems for research ideation \citep{wang-etal-2024-scimon, si2025can, garikaparthi-etal-2025-iris, gottweis2025aicoscientist, baek-etal-2025-researchagent}. Although AI-generated ideas score higher than human ideas on criteria such as human-evaluated novelty at the \textit{ideation} stage  \citep{si2025can}, 
many of these seemingly promising ideas turn out to be ineffective when \textit{executed} \citep{si2025ideationexecutiongapexecutionoutcomes}. 
The execution of faulty ideas can lead to substantial costs, particularly in fields requiring significant computational resources or wet-lab experiments. There is thus a critical need to rigorously evaluate research ideas pre-execution to prioritize the strongest ones for resource investment. 


Despite this need, automatic research idea evaluation remains an underexplored area. Some works narrowly frame it as a prediction task: deciding which idea among a pair would lead to better results on predefined benchmarks \citep{wen2025predictingempiricalairesearch}. Others focus on one-dimensional approaches to idea evaluation (e.g., novelty) \citep{afzal2025notnovelenoughenriching, shahid-etal-2025-literature},
or target only specific sub-disciplines (e.g., AI) \citep{si2025can}. Importantly, these systems either only generate scores or are limited to sparse feedback such as short rationale statements \citep{feng2025grapheval, wen2025predictingempiricalairesearch, shahid-etal-2025-literature}. However, to create AI co-scientists that can generate and refine research ideas, giving dense, actionable, and multifaceted feedback is crucial \citep{wu2023finegrained, cao-etal-2024-enhancing}. 
To the best of our knowledge, no existing work addresses the challenge of comprehensive research idea evaluation across disciplines within a framework that provides detailed and actionable feedback for idea refinement.

We introduce \system (\autoref{fig:highlight}), a 
research idea evaluation system
grounded in the most recent literature. \system evaluates research ideas based on two fundamental criteria: \textit{soundness} and \textit{contribution}. \textbf{(1) Soundness} refers to the empirical validity of each proposed method in the research plan, assessed by systematically examining whether similar applications of this method in existing literature have demonstrated success or failure. 
\textbf{(2) Contribution} refers to the degree of advancement a research idea offers across different dimensions--e.g., its proposed methodology, data, evaluation approaches, and conceptual framework--relative to existing literature. 
The rationale for dimension-based evaluation is that novelty is multi-faceted by nature, and an idea can be considered novel relative to certain aspects of prior work, rather than being categorically novel or not \citep{rubaiat2025mappingevolutionresearchcontributions, radensky2025scideatorhumanllmscientificidea}.
Given a research idea detailing the problem, proposed methodology, and planned experiments, \system employs a multi-stage pipeline that first generates targeted search queries to retrieve a large volume of related literature from Semantic Scholar \citep{kinney2025semanticscholaropendata} (\autoref{fig:scholareval}). It then extracts key information from the retrieved literature to assess the research plan's soundness and contribution. Finally, it synthesizes detailed feedback supported by relevant citations. 

To evaluate \system, we construct \benchmark, a multi-disciplinary dataset of 117 research ideas and their corresponding reviews validated by subject-matter experts across four disciplines: artificial intelligence, biochemistry, neuroscience, and ecology. As showcased in \autoref{fig:highlight}, reviews in \benchmark are composed of multiple rubrics, each focusing on a specific point that the evaluation should address, for 1076 rubrics in total. 
We develop a multi-faceted automatic evaluation framework to assess \system against strong baselines, namely state-of-the-art LLMs and deep research systems. 
Our results show that \system achieves greater coverage of the expert-annotated review rubrics in \benchmark, significantly outperforming all baselines and surpassing \hnm{OpenAI Deep Research}
by over 20\% relative improvement. Our results also demonstrate that \system is consistently preferred over \hnm{OpenAI} Deep Research in terms of evidence support, depth, and actionability.

Our human study involving 18 experts and 46 evaluations further supports the real world usefulness of \system across our four target disciplines. \system outperforms deep research by a significant margin in metrics tied to our core contributions: literature engagement, citation use, feedback validity, idea refinement, relevant evaluation aspects, and overall usefulness. 
%
Our work makes the following major contributions: 
\begin{itemize}[leftmargin=*]
    \item \system, a literature-grounded framework that comprehensively evaluates research ideas based on their soundness and contribution with actionable feedback. We will openly release the \system tool and user interface.  
    \item \benchmark, an expert-annotated dataset for research idea evaluation spanning four disciplines and composed of 117 research ideas with 1076 detailed review rubrics.
    \item A multifaceted evaluation for long-form research idea review responses, including automatic metrics and a carefully designed human expert evaluation.
\end{itemize}
\section{\system}
\label{sec:scholareval}
\system is a retrieval-augmented, multi-stage pipeline designed to give an in-depth evaluation of research ideas based on their soundness and contribution. 

\textbf{Task Formulation.} Given a research idea $I$, 
the task is to find papers $P = \{p_1, p_2, p_3, \ldots\}$ that are highly relevant to $I$ and synthesize their findings in the context of $I$ to generate a comprehensive evaluation $\mathcal{E} = (S, C)$ covering its soundness and contribution. The evaluation is accompanied by citations, ensuring that all claims are supported by evidence from the literature.

\begin{figure*}[t]  
    \centering
    \includegraphics[width=1\linewidth]{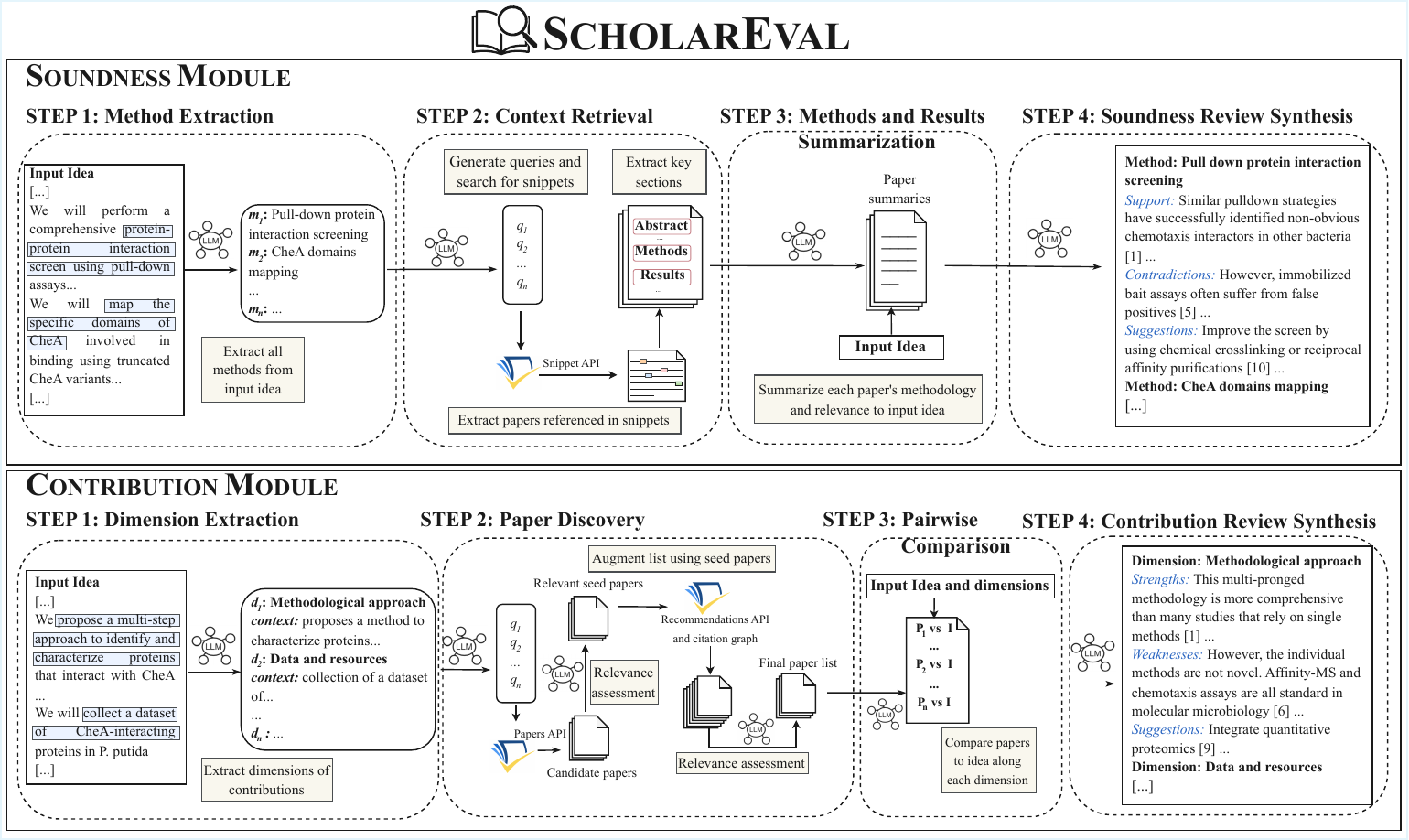}
    \caption{Overview of \system and its two modules. \textbf{Top:} The soundness module extracts the methods proposed in the research idea and conducts a thorough literature search for similar applications of each method to determine its potential effectiveness. \textbf{Bottom:} The contribution module identifies the dimensions along which the research idea is making contributions and conducts detailed comparisons with related papers along each dimension to identify areas of novelty or lack thereof.}
    \label{fig:scholareval}
\end{figure*}

\textbf{Overview of \system.} As shown in \autoref{fig:scholareval}, \system is composed of two main modules: \textbf{Soundness} and \textbf{Contribution}. We present both their workflows in \Sref{sec:soundness} and \Sref{sec:contribution} and provide further details in \autoref{app:system}.
\subsection{Soundness Evaluation}
\label{sec:soundness}
The soundness module evaluates the methodological rigor of an idea by extracting methods, searching for relevant literature, and synthesizing evidence to assess whether the proposed approaches are well-supported or contradicted by existing work. 

\textbf{Method Extraction.} The first objective of the soundness pipeline is to extract distinct methodological components including algorithmic approaches, experimental designs, evaluation protocols, ablation studies, or analytical frameworks from the idea. Formally, we leverage an LLM, referred to hereafter as \( \mathcal{M} \), to extract all methods $M = \{m_1, m_2, \ldots, m_k\}$ from the research idea $I$.  

\textbf{Context Retrieval.} This step gathers information from the literature about the effectiveness of the proposed methods. For each extracted method $m_i \in M$, \( \mathcal{M} \) generates a relevant query for \href{https://api.semanticscholar.org/api-docs/#tag/Snippet-Text/operation/get_snippet_search}{Semantic Scholar snippet search}  \citep{kinney2025semanticscholaropendata}, which indexes 285.6M passages extracted from the title, abstract, or body of research papers \citep{singh2025ai2scholarqaorganized}. 
Since the queries are constructed from the description of the method $m_i$, it is likely to retrieve snippets within relevant methodology sections. We extract all papers referenced in these snippets, which provides a dense collection of relevant sources to broaden the understanding of $m_i$. 
We download the full text of these papers and parse them using \href{https://github.com/kermitt2/grobid}{GROBID}, a state-of-the-art document parsing tool, to extract three key sections from each paper: the methods section, to compare its similarity with the current method $m_i$; the results section, to judge the method's effectiveness in a given context; and the abstract for a holistic view of the paper. At the end of this process, we obtain a list of related papers $P_i = \{p_{i,1}, p_{i,2}, \ldots, p_{i,n_i}\}$ for each method $m_i$, where each paper is represented as a triplet \texttt{(abstract, methods, results)}. This list constitutes essential literature context to evaluate each method's effectiveness. 

\textbf{Methods and Results Summarization.} This stage serves two key functions: first, it filters out extracted papers that are not relevant to assessing the method $m_i$, and second, it condenses the most vital information from relevant papers, since retaining all paper data results in prohibitive context lengths. Specifically, for each paper $p_{i,j} \in P_i$, we instruct \( \mathcal{M} \) to identify whether the methodology described in the paper is relevant to the method $m_i$, and if so, generate a compact summary of its methods and results grounded in the context of the method $m_i$ and the overall research idea $I$. 

\textbf{Soundness Review Synthesis.} Grounded in the condensed context, the soundness pipeline concludes by synthesizing method-level soundness evaluations. Specifically, \( \mathcal{M} \) analyzes the paper summaries in the context of the method $m_i$ and the research idea $I$ to synthesize three main sections: (1) \textit{Support}: the support for the method $m_i$ based on the literature. This section details whether there are similar methods in the literature that have shown successful results, and uncovers how they relate to the current $m_i$. (2) \textit{Contradictions}: the contradictions to the method $m_i$ based on the evidence extracted from the literature. In relation to the proposed method $m_i$, this section highlights the limitations that methods in a similar context have faced, signaling
its potential ineffectiveness. (3) \textit{Suggestions}: based on the strengths and limitations of the method identified in the previous sections, \( \mathcal{M} \) generates actionable suggestions for improvement to refine the proposed methodology. 

\subsection{Contribution Evaluation}
\label{sec:contribution}
The contribution module 
assesses the novelty and significance of an idea by identifying contribution dimensions, discovering related papers, conducting pairwise comparisons to determine how the idea advances the literature, and synthesizing a dimension-level contribution review.

\textbf{Dimension Extraction.} The initial step of contribution evaluation consists of extracting the dimensions along which the research idea is making contributions to the field. Dimensions represent the facets of the idea's potential contributions that are specific and comparable across related literature (e.g., system design, data collection, evaluation methodology, etc.). Instead of imposing pre-defined dimensions as in \citet{radensky2025scideatorhumanllmscientificidea}, we use LLM extraction to ensure flexibility based on the nature of the research idea \citep{rubaiat2025mappingevolutionresearchcontributions}. Specifically, given the research idea $I$, we use \( \mathcal{M} \) to extract dimensions $D = \{d_1, d_2, \ldots, d_l\}$. Examples of dimensions include tool or system design, conceptual framework, evaluation methodology, etc.
Each $d_i$ also 
includes the reasoning for how the idea makes contributions along that dimension. These statements will provide important context to ground the query generation
in the subsequent step. 

\textbf{Paper Discovery.} In this step, we conduct a broad search over the literature to identify relevant papers to compare $I$ against. However, unlike the soundness module, which requires searching paper content for methodological details, contribution evaluation can be performed using only abstracts, since a paper's main contributions are typically highlighted there. This also allows us to cast a wider net and gather a broad set of related papers. Such breadth is essential for contribution evaluation, as determining truly novel contributions requires an exhaustive view of the literature.
The paper discovery thus consists of the following steps: (1) For each extracted dimension, \( \mathcal{M} \) generates queries to search for relevant papers using \href{https://api.semanticscholar.org/api-docs/#tag/Paper-Data/operation/get_graph_paper_relevance_search}{Semantic Scholar paper search} and retrieve their abstracts. (2) \( \mathcal{M} \) assesses similarity in contributions of each candidate paper abstract relative to the research idea $I$ and assigns a score on a scale from 1 to 5. Papers that are deemed highly relevant (i.e., score $\geq 3$) are then used as seeds for the paper augmentation stage. (3) Paper augmentation leverages the \href{https://api.semanticscholar.org/api-docs/recommendations}{Semantic Scholar Recommendations API} to find similar papers, and we additionally extract the publications cited by each seed paper. (4) This augmented list of candidate papers then undergoes another stage of relevance assessment by \( \mathcal{M} \). However, due to the typically large volume of papers at this stage, we first filter this list to the top $n$ papers based on semantic embedding\footnote{We use Titan Text Embedding v2 \citep{aws_titan_embeddings}} similarity between abstracts and $I$ before forwarding to \( \mathcal{M} \).

\textbf{Pairwise Comparison.} Once the final set of papers $P_D = \{p_1, p_2, \ldots, p_m\}$ is identified, we conduct a series of pairwise comparisons to uncover how these papers' contributions compare to those proposed in $I$. Specifically, we prompt \( \mathcal{M} \) to compare each paper's abstract to $I$ along each dimension $d_i \in D$. This comparison produces a granular view of the areas in which $I$ is making novel advancements and those in which it is lacking in novelty, compared to existing work. 

\textbf{Contribution Review Synthesis.} The final stage of contribution evaluation consists of synthesizing dimension-level contribution assessments. For each $d_i \in D$, \( \mathcal{M} \) uses the results of the pairwise comparisons to synthesize an evaluation composed of three sections: (1) \textit{Strengths}: the novel contributions that $I$ makes along this dimension compared to prior work. (2) \textit{Weaknesses}: the areas in which the contributions of $I$ are lacking or limited compared to prior work. (3) \textit{Suggestions}: actionable recommendations to improve the novelty of $I$ along this dimension. 


\section{\benchmark: A dataset for research idea evaluation}
\label{sec:benchmark}
\begin{figure*}[t]  
    \centering
    \includegraphics[width=1\linewidth]{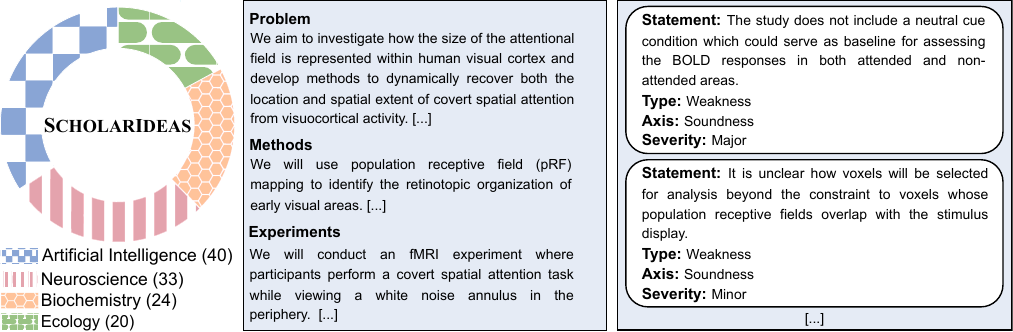}
    \caption{Overview of \benchmark. \textbf{Left:} \textsc{ScholarIdeas} includes 117 research ideas across four disciplines. \textbf{Right:} Example of a neuroscience research idea and review in \textsc{ScholarIdeas}. Each review is composed of multiple rubrics, for a total of 1076 rubrics across the dataset. This idea-review pair is adapted from \citep{Bloem2025Dynamic}.}
    \label{fig:scholar_ideas}
\end{figure*}
We design \system to provide a holistic evaluation of research ideas within a domain-agnostic framework.  
A meaningful assessment of \system's quality thus hinges on a multi-domain collection of research ideas paired with ground-truth reviews.
However, existing datasets and benchmarks fall short: they focus on full paper reviews rather than research ideas \citep{kang-etal-2018-dataset, weng2025cycleresearcher}, capture singular dimensions such as novelty \citep{shahid-etal-2025-literature}, or are restricted to one discipline \citep[e.g., AI;][]{si2025can}. 

To this end, we curate \benchmark, a dataset containing research ideas and reviews covering four disciplines: artificial intelligence, neuroscience, biochemistry, and ecology. \hnm{Due to a lack of sources with pre-execution research ideas, we employ a semi-automatic pipeline to retrospectively extract ideas from existing papers, where each example undergoes rigorous evaluation by subject-matter experts (\Sref{sec:data-curation})}. Our primary motivation for developing \system is to evaluate AI-generated ideas. However, collecting detailed expert reviews for AI-generated ideas at scale is prohibitively expensive and practically infeasible. Our dataset of existing human-written ideas with expert reviews serves as a reliable proxy. 

The evaluation of long-form responses, such as the ones generated by \system, is also inherently challenging. While other scientific tasks admit easily verifiable success criteria (e.g., code generation \citep{jansen-etal-2025-codescientist, chen2025scienceagentbench, li2025autosdt}), such evaluation metrics are not readily available for our use case. 
We develop 
a multi-faceted automatic evaluation pipeline detailed in \Sref{sec:metrics}. We further corroborate our automatic evaluation with a large-scale human evaluation (\Sref{sec:expert-eval}).

\subsection{Data Curation and Annotation}
\label{sec:data-curation}

\textbf{Data selection.} We manually select papers and reviews from two sources: \href{https://openreview.net/group?id=ICLR.cc/2025/Conference}{OpenReview (ICLR 2025)} for AI-related research ideas and \href{https://elifesciences.org/}{eLife} for life sciences. Both sources include high-quality reviews by multiple reviewers for all submitted papers, 
in contrast to many other sources that do not openly release reviews or restrict them to those of accepted manuscripts. To ensure the suitability of the reviews to be used as ground-truth, we only select papers satisfying the following criteria: (1) The paper must be reviewed by at least two reviewers. (2) All reviews of the paper are in agreement and offer a general consensus about the quality of the work. (3) The reviews mention criticism about the underlying research idea and not exclusively about obtained results or other details known post-execution. For each of the sources we use, we only retrieve the first version of the submission prior to any revisions, along with the first round of reviews. We also annotate each idea with its publication date, which we use as a cutoff date for literature search during \system and baseline execution. 

\textbf{LLM-based extraction.} After identifying 130 papers across the four disciplines of interest satisfying our criteria, we follow a multi-step approach to extract the research idea and its corresponding reviews. \textbf{(1) Document parsing.} We first use GROBID to parse the paper into different sections and employ a heuristic on section names to only keep those related to the background and methodology, while dropping all sections mentioning experiment execution and results. \textbf{(2) Research idea extraction.} We instruct an LLM\footnote{We use Claude 4 Sonnet} to extract the research idea based on the parsed documents such that each research idea is composed of three components (Problem, Methods, Experiments). \textbf{(3) Review rubrics extraction.} Finally, we provide an LLM with the full reviews corresponding to the paper along with the extracted research idea and instruct it to extract statements from the reviews pertaining to assessments of the underlying research idea.
These statements are further classified by the LLM based on their type (strength or weakness), severity (major or minor), and axis (soundness or contribution) to construct our review rubrics.

\textbf{Expert Validation.} We invite 6 subject-matter experts (PhD students, postdocs, and professors) in artificial intelligence, neuroscience, biochemistry, and ecology to validate the quality of each \texttt{(research idea, review rubrics)} pair extracted by the LLM. Namely, the experts are instructed to ensure that the research idea is a faithful representation of the paper at the ideation stage, that the review rubrics do not mention execution, results, or paper presentation, and that the research idea and review rubrics are consistent (i.e. the review only addresses aspects contained in the research idea). Annotators are also asked to verify the correctness of the dimension assignments made by the LLMs. They are instructed to make necessary corrections and to discard instances not in their area of expertise.  

At the end, we obtain 117 validated \texttt{(research idea, review rubrics)} pairs balanced across four disciplines. \autoref{fig:scholar_ideas} shows the statistics of \benchmark and an example pair. Further details about \benchmark curation are given in \autoref{app:benchmark}. 

\subsection{Evaluation Protocol}
\label{sec:metrics}
\begin{table*}[t]
\small
\centering
\begin{tabular}{lccccc}
\toprule
\textbf{} & \textbf{Overall} & \textbf{AI} & \textbf{Neuroscience} & \textbf{Biochemistry} & \textbf{Ecology} \\
\midrule
\textcolor{gray}{\textit{Language Models}} & &  &  &  & \\
Llama-3.3-70B & 1.83 $\pm$ 0.89 & 1.88 $\pm$ 0.92 & 1.82 $\pm$ 0.86 & 1.81 $\pm$ 0.90 & 1.74 $\pm$ 0.87 \\
GPT-4.1 & 2.18 $\pm$ 1.08 & 2.21 $\pm$ 1.12 & 2.18 $\pm$ 1.10 & 2.14 $\pm$ 1.04 & 2.13 $\pm$ 1.00 \\
GPT-5.1 Instant & 1.94 $\pm$ 0.98 & 2.09 $\pm$ 0.99 & 1.85 $\pm$ 0.99 & 1.68 $\pm$ 0.91 & 1.95 $\pm$ 0.97 \\
Claude-4-Sonnet & 2.18 $\pm$ 1.04 & 2.28 $\pm$ 1.02 & 2.15 $\pm$ 1.08 & 2.01 $\pm$ 1.01 & 2.11 $\pm$ 1.03 \\
\midrule
\textcolor{gray}{\textit{Web-connected LM}} & &  &  &  & \\
GPT-4o-search-preview & 1.90 $\pm$ 0.98 & 1.95 $\pm$ 0.97 & 1.84 $\pm$ 0.96 & 1.86 $\pm$ 1.00 & 1.95 $\pm$ 1.02 \\
\midrule
\textcolor{gray}{\textit{Deep Research}} & &  &  &  & \\
OpenAI Deep Research & 2.28 $\pm$ 1.07 & 2.35 $\pm$ 1.07 & 2.25 $\pm$ 1.04 & 2.08 $\pm$ 1.06 & 2.35 $\pm$ 1.11 \\
DR Tulu & 2.33 $\pm$ 1.14 & 2.35 $\pm$ 1.11 & 2.31 $\pm$ 1.17 & 2.24 $\pm$ 1.04 & 2.39 $\pm$ 1.20 \\
\midrule
\textcolor{gray}{\textit{Ours}} & &  &  &  & \\
\system$_{\text{Llama}}$ & 2.04 $\pm$ 1.16$^{*}$ & 2.06 $\pm$ 1.14$^{*}$ & 2.05 $\pm$ 1.18$^{*}$ & 2.04 $\pm$ 1.19$^{*}$ & 1.94 $\pm$ 1.13$^{*}$ \\
\system$_{\text{GPT-4.1}}$ & 2.72 $\pm$ 1.47$^{\dagger}$ & 2.84 $\pm$ 1.39$^{\dagger}$ & \textbf{2.74 $\pm$ 1.55$^{\dagger}$} & 2.61 $\pm$ 1.42$^{\dagger}$ & 2.52 $\pm$ 1.51$^{\dagger}$ \\
\system$_{\text{GPT-5.1}}$ & 2.17 $\pm$ 1.27$^{*}$ & 2.31 $\pm$ 1.30$^{*}$ & 2.12 $\pm$ 1.24$^{*}$ & 1.96 $\pm$ 1.19$^{*}$ & 2.09 $\pm$ 1.28$^{*}$\\
\system$_{\text{Claude}}$ & \textbf{2.77 $\pm$ 1.40$^{\dagger}$} & \textbf{2.91 $\pm$ 1.34$^{\dagger}$} & 2.55 $\pm$ 1.45$^{\dagger}$ & \textbf{2.64 $\pm$ 1.40$^{\dagger}$} & \textbf{2.90 $\pm$ 1.39$^{\dagger}$} \\
\bottomrule
\end{tabular}
\caption{\metricpillblue{\small Coverage $\uparrow$} of baselines and variants of \textsc{ScholarEval} overall and per-discipline. $^{*}$  indicates significant improvement over at least one baseline and $^{\dagger}$ indicates significant improvement over all baselines. Best results are in bold. Statistical significance details are provided in \autoref{app:results}.}
\label{tab:main_results}
\end{table*} 
A high-quality research idea evaluation should highlight all strengths and weaknesses that an expert reviewer would mention, ground its claims in relevant literature, engage deeply with the points discussed and the works cited, and go beyond simple good or bad judgment to offer actionable suggestions for improvement. Guided by this desiderata, we develop automatic metrics to assess the performance of \system and baselines. Full details of prompts and implementation are provided in the Appendix \autoref{app:metrics}.  

\textbf{\metricpillblue{\small Coverage  $\uparrow$}.} 
This is a recall-based metric which measures the extent to which an evaluation covers the rubrics from the corresponding review in \benchmark. \hnm{Specifically, we use \href{https://github.com/prometheus-eval/prometheus-eval}{Prometheus-Eval} \citep{kim2024prometheus} with a GPT-4 backbone \citep{openai2024gpt4technicalreport}, a framework that is widely used to evaluate long-form output and that is shown to align well with human judgment \citep{asai2024openscholarsynthesizingscientificliterature, yifei2025researchqaevaluatingscholarlyquestion, shao-etal-2024-assisting, bonomo-etal-2025-literaryqa}}. We instruct it to assign a 1–5 score based on how well each evaluation addresses the points in the reference rubric. The final score is the average over all 1,076 rubrics in \benchmark.

\textbf{\metricpillblue{\small Reference Inv. $\downarrow$}.} To compute the reference invalidity rate, we automatically check the references (i.e., paper links) cited in the evaluation to verify whether they correspond to existing papers. This is done by examining the status code returned for each link. Since edge cases arise where status codes do not clearly indicate validity, we conservatively report a lower bound: the proportion of references that are clearly invalid. 

\textbf{\metricpillblue{\small Evidence}, \metricpillblue{\small Act}, \metricpillblue{\small Depth}.} We use an LLM-as-judge, Claude 4 Sonnet, to give pairwise preferences between \system and our strongest baseline along three criteria: \textbf{Evidence Support}, which measures how well claims are grounded in literature and supported by relevant citations; \textbf{Actionability}, which captures the clarity, usefulness, and feasibility of the suggestions; and \textbf{Depth}, which evaluates the level of engagement with each point and whether the evaluation mentions specifics about the work it cites rather than relying on generic statements. 
\hnm{We also compute agreement with human annotators, highlighting strong alignment between the LLM judge and human judgment (see \autoref{app:metrics})}. 

\section{Experiments and Results}
\label{sec:ex-results}
\subsection{Experimental Details}
\label{sec:ex-details}
\textbf{Baselines.} 
\hnm{We compare \system{} against baselines that are capable of providing detailed research idea feedback at a similar scope. Specifically, we select a host of strong baselines consisting of frontier open-source and closed-source LLMs—Llama-3.3-70B \citep{grattafiori2024llama3herdmodels}, GPT-4.1 \citep{openai2025introducing}, GPT-5.1 Instant \citep{openai2025gpt5.1}, Claude-4-Sonnet \citep{anthropic2025claude}, and GPT-4o-search-preview \citep{openai2025searchpreview} as a web-connected LLM baseline—in addition to two deep research systems: OpenAI Deep Research\footnote{o4-mini-deep-research} \citep{openai2025introducingdeep}, a strong and widely used deep research system and the only proprietary one available via API at the time of writing, and DR-Tulu \citep{shao2025dr}, a recent open-source deep research system that rivals proprietary ones and shows strong performance for academic tasks.}

\hnm{While there exist related academic systems for research idea evaluation, these primarily focus on idea scoring or ranking \citep{feng2025grapheval, wen2025predictingempiricalairesearch}, or provide only short rationales \citep{shahid-etal-2025-literature}, and thus do not offer feedback at the same scope as \system. We therefore do not treat them as primary baselines, but provide a small-scale comparison in \autoref{app:results} for context.}

We prompt baselines with a detailed template aligned with \system output format and depth, including method and dimension decomposition and dedicated sections for strengths, weaknesses, and suggestions. 

\textbf{Instantiation.} We evaluate four variants of our system: \system{}$_{\text{Llama}}$, \system{}$_{\text{GPT-4.1}}$, \hnm{\system{}$_{\text{GPT-5.1}}$}, and \system{}$_{\text{Claude}}$, which respectively use Llama-3.3 70B, GPT-4.1, GPT-5.1 Instant, and Claude 4 Sonnet as backbones. 

Full prompts and additional experimental setup details are provided in \autoref{app:exp}.

\subsection{Results and Analysis}
\label{sec:results}
\paragraph{\system outperforms baselines in rubric coverage across disciplines.}
As shown in \autoref{tab:main_results}, 
\system{}$_{\text{GPT-4.1}}$ and \system{}$_{\text{Claude}}$ achieve statistically significant gains on \metricpillblue{\small Coverage  $\uparrow$} over all baselines overall and in every discipline. \system{}$_{\text{Llama}}$ \hnm{and \system{}$_{\text{GPT-5.1}}$ also surpass their respective backbone models}, indicating that \system delivers improvements across models. Importantly, these gains are not explained by emphasizing minor rubric items only; \system also provides better coverage of expert-annotated \emph{major} points (see \autoref{tab:type_axis_severity} in Appendix).


\textbf{\system eliminates reference invalidity.}
\hnm{As shown in \autoref{tab:ref-invalid}, \system variants and DR-Tulu achieve 0\% \metricpillblue{\small Reference Inv. $\downarrow$} in our evaluation, while OpenAI Deep Research exhibits a similarly low invalidity rate of 1\%, reflecting that information attribution has become increasingly robust in modern research-oriented systems.} As previously stated, baseline rates represent lower bounds. However, a manual audit of a sample of outputs (see \autoref{app:manual-audit}) indicates that up to 80\% of Llama-3.3-70B citations are hallucinated in certain evaluations. Our audit also uncovers subtler failures, including misattributed authors and claims that are inconsistent with the cited sources. These patterns align with findings from recent evaluations of deep research systems \citep{li2025reportbenchevaluatingdeepresearch}. In contrast, \system eliminates these issues entirely, ensuring that all citations resolve to valid, traceable sources.

\textbf{Quality of \system evaluations exceeds that of deep research \hnm{systems}.} We compare \system{}$_{\text{Claude}}$ to the two strong baselines, \hnm{OpenAI Deep Research and DR Tulu} based on the metrics \metricpillblue{\small Evidence}, \metricpillblue{\small Act} and \metricpillblue{\small Depth}. Results in \autoref{tab:pairwise-comparison} indicate that \system$_{\text{Claude}}$ is consistently better \hnm{than both} at giving evidence-based evaluations, making actionable suggestions to improve the research idea, as well as giving sufficient details and deeply engaging with the literature it cites. Our user study detailed in \Sref{sec:expert-eval} further corroborates these results by the preference of human users.
\begin{table}[t]
  \centering
  \small
  \begin{tabular}{lc}
    \toprule
    \textbf{} & \textbf{\metricpillblue{\small Reference Inv. $\downarrow$}} \\
    \midrule
    Llama-3.3-70B & 19.07\% \\
    GPT-4.1       & 15.22\% \\
    \hnm{GPT-5.1 Instant}       & \hnm{23.33}\% \\
    Claude-4-Sonnet & 13.90\% \\
    GPT-4o-search-preview & 1.66\% \\
    OpenAI Deep Research & 1.07\% \\
    \hnm{DR Tulu} & \hnm{0}\% \\
    \midrule[0.75pt]
    \system$_{\text{Llama}}$  & 0\% \\
    \system$_{\text{GPT-4.1}}$    & 0\% \\
    \hnm{\system$_{\text{GPT-5.1}}$}    & 0\% \\
    \system$_{\text{Claude}}$ & 0\% \\
    \bottomrule
  \end{tabular}
  \caption{Rate of reference invalidity across all systems. Baseline values are lower bounds; actual invalidity is higher, especially for non-retrieval systems. Reference invalidity is not an issue in \system.}
  \label{tab:ref-invalid}
\end{table}

\begin{table}[t]
\centering
\small
\begin{tabular}{lcc}
\toprule
\textbf{Metric} 
& \textbf{vs. OpenAI DR} 
& \textbf{vs. DR-Tulu} \\
\midrule
\metricpillblue{\small \textbf{Evidence}}
&  76.6\%
&  64.0\% \\
\metricpillblue{\small \textbf{Act}} 
&  71.2\%
&  62.2\% \\
\metricpillblue{\small \textbf{Depth}} 
&  83.8\%
&  66.7\% \\
\bottomrule
\end{tabular}
\caption{Pairwise win rate between \system$_{\text{Claude}}$ (our best performing system) and OpenAI Deep Research and DR Tulu using LLM judge.}
\label{tab:pairwise-comparison}
\end{table}

\textbf{Ablations studies.} We conduct ablations to assess the effectiveness of individual components of \system. Namely, we remove each of the methods and results extraction (MRE), paper augmentation (PA), and pairwise comparison (PC). Details about the setup for each ablation experiment can be found in \autoref{app:exp}. Results in  \autoref{tab:ablations} indicate that the removal of each of these components leads to degradation in performance of \system on \metricpillblue{\small Coverage $\uparrow$}. This is likely due to the reduced information given to the model (i.e. the effectiveness of the methods from prior work based on the reported results, the dense list of relevant papers, and the granular dimension-level paper comparisons). These results showcase the importance of these steps in providing \system with essential context from the literature to generate effective idea evaluations.
\begin{table}[t]
\centering
\small
\begin{tabular}{l c}
\toprule
\textbf{} & \textbf{\metricpillblue{\small Coverage $\uparrow$}} \\
\midrule
\system$_{\text{Claude}}$ & 2.91 $\pm$ 1.34 \\
\quad -MRE  & 2.47 $\pm$ 1.42 \\
\quad -PA   & 2.42 $\pm$ 1.28 \\
\quad -PC   & 2.39 $\pm$ 1.23 \\
\bottomrule
\end{tabular}
\caption{Ablations of different components of \system on \textsc{ScholarIdeas-AI}.}
\label{tab:ablations}
\end{table}
\section{Expert User Study}
\label{sec:expert-eval}

\textbf{Design.} We bolster our automatic evaluation with a large-scale blinded user study to further gauge the real-world usefulness of \system. We recruit 18 experts for 46 total evaluations in a blind experiment with \hnm{OpenAI Deep Research} \cite{openai2025introducingdeep}. 
Each expert had an education level of PhD student or beyond and was verified to have at least one paper published in their field (see \autoref{tab:user_study_demographics}). We create a blind interface for experts to interact with \system or OpenAI Deep Research (see \autoref{fig:us_ui} and \autoref{fig:us_ui2}). We choose OpenAI DeepResearch as it was both the strongest and most popular model for our task at the time of conducting this experiment.

\textbf{Procedure.} We ask each expert participant to first write a research idea that includes the problem they aim to tackle and the suggested methodology. This can be one they have yet to experiment with, or an idea they have already published. 
After receiving the soundness and contribution evaluation from a single system (either \system or OpenAI deep research), they complete a detailed rubric related to the core components of idea evaluation (e.g. the usefulness of the suggestions, the engagement with literature, etc.). 
The entire process for creating an idea, generating the feedback, and scoring the rubric took an average of 1 hour across participants, and we allow up to 4 unique idea submissions. The exact questions, user demographics, blindness and randomization validity, and compensation details can be found in \autoref{app:userstudy}.

\textbf{Evaluation.}
We group questions into six dimensions. \metricpillorange{\small Citations} captures the number of references experts would actually use in their research. Because research ideas can be complex and domain-specific, we use  \metricpillorange{\small Faithful} to measure the extent to which the feedback aligns with nuances of the research idea. \metricpillorange{\small Useful} reflects the helpfulness of each system and the expert's enthusiasm for future use. \metricpillorange{\small Focus} measures the degree to which each system targets the most important factors of each research idea. \metricpillorange{\small LitEngage} indicates the depth each system uses when making detailed comparisons with specific components of relevant literature. \metricpillorange{\small Refine} gauges whether the system provides valuable, targeted, feasible suggestions for improvement that experts believe would actually improve their research idea. The questions related to each dimension are available in \autoref{tab:user_study_questions}. We refer readers to \ref{app:user_study_stats} for details on linear mixed effects model, which we adopt as our statistical method.
\textbf{Results and analysis.} Results in \autoref{tab:user_study} show a statistically significant preference for \system over \hnm{OpenAI Deep Research} across all six dimensions measured in our expert user study. Experts found 1.5 more useful \metricpillorange{\small Citations} using \system and scored \metricpillorange{\small LitEngage} 1.2 higher, demonstrating the effectiveness of our multi-stage literature retrieval pipeline. A strong effect on \metricpillorange{\small Faithful} and \metricpillorange{\small Focus} underscores the benefit of systemically breaking down the research plan into smaller, controlled comparisons with relevant literature. The actionable feedback generated by \system had clear advantages as well, with experts reporting a 1.2 increase in \metricpillorange{\small Refine}. The higher score on \metricpillorange{\small Useful} further underscores the overall usefulness of \system as research evaluation framework as judged by experts.

\section{Related Work}
\label{app:relatedwork}

\textbf{Literature Grounded Systems for Research.} Multiple works have been proposed that use LLMs in literature grounded systems to assist in research. A commonly targeted use case for these systems is literature synthesis and related work generation \citep{asai2024openscholarsynthesizingscientificliterature, kangetalsynergi, agarwal2025litllmtoolkitscientificliterature, ding2025sciragadaptivecitationawareoutlineguided}. In addition, other systems have been developed for literature understanding and question answering \citep{skarlinski2024languageagentsachievesuperhuman, singh2025ai2scholarqaorganized, volkova2025crossdisciplinaryknowledgeretrievalsynthesis}. \system builds on techniques inspired by these works but focuses on the underexplored problem of literature-grounded research idea evaluation. 

\textbf{Research Ideation.} Recently, there has been growing interest in using LLMs for generating research ideas and scientific hypotheses, often from a literature-driven perspective. For example, \citet{wang-etal-2024-scimon} proposed SciMON, a framework that retrieves inspirations from past scientific papers to generate novel ideas. Similarly, \citet{baek-etal-2025-researchagent} employ an iterative approach over related papers and a knowledge store to generate scientific hypotheses. Other works emphasize human-LLM collaboration for idea generation \citep{radensky2025scideatorhumanllmscientificidea, garikaparthi-etal-2025-iris, puetalideasynth, shao2025omniscientistcoevolvingecosystemhuman, liu2025perspectrachoosingexpertsenhances}, while multi-agent approaches to generate and refine hypotheses increase in popularity \citep{gottweis2025aicoscientist, ueda2025exploringdesignmultiagentllm, mitchener2025kosmosaiscientistautonomous}. Furthermore, \citet{si2025can} conducted a large-scale study assessing both human- and LLM-generated ideas in the AI field, finding that LLMs are capable of generating more novel ideas based on human evaluation. 

\textbf{Research Idea Evaluation.} Many of the aforementioned ideation systems also incorporate modules for idea review and refinement \citep{baek-etal-2025-researchagent, radensky2025scideatorhumanllmscientificidea}. For instance, \citet{baek-etal-2025-researchagent} use a review agent that is prompted with a rubric induced from human preference. Similarly, \citet{shahid-etal-2025-literature, radensky2025scideatorhumanllmscientificidea} introduce a retrieval-augmented framework that generates novelty classifications with brief reasoning. Additionally, \citet{feng2025grapheval} introduced a graph-based LLM framework to score research ideas. A concurrent work by \citet{afzal2025notnovelenoughenriching} proposes an LLM framework specifically for paper novelty evaluation. \citet{wen2025predictingempiricalairesearch} tackle the problem of idea evaluation by building a system that predicts empirical outcomes in AI research, choosing the most promising idea among a pair. 
\system addresses many of the limitations of these systems. First, rather than relying on the parametric knowledge of LLMs for idea evaluation as in \citet{baek-etal-2025-researchagent}, \system incorporates a carefully designed retrieval pipeline to ground evaluation in the most recent literature. Second, it goes beyond one-dimensional evaluation (e.g., novelty) \citep{shahid-etal-2025-literature, afzal2025notnovelenoughenriching} to also assess the validity of proposed methods, since novelty alone is not a sufficient condition for successful research execution \citep{si2025ideationexecutiongapexecutionoutcomes}. Finally, \system places strong emphasis on generating dense and actionable feedback for idea refinement, in contrast to systems that only produce scores \citep{feng2025grapheval, wen2025predictingempiricalairesearch} or sparse feedback \citep{shahid-etal-2025-literature}.

\newcommand{\sym}[1]{\ensuremath{^{#1}}} 
\begin{table}[t]
\centering
\small
\begin{tabular}{l
                l 
                c                                              
                c}                                             
\toprule
Dimension & \phantom{0}{$\beta$} & {\system} & {OpenAI DR} \\
\midrule
\metricpillorange{\small Citations}    & 0.81\textsuperscript{***} & 2.66 $\pm$ 2.43 & 1.09 $\pm$ 1.28 \\
\metricpillorange{\small Faithful}     & 0.69\textsuperscript{*}   & 3.88 $\pm$ 1.50 & 2.91 $\pm$ 1.19 \\
\metricpillorange{\small Useful}       & 0.63\textsuperscript{*}    & 7.80 $\pm$ 1.71 & 6.56 $\pm$ 2.05 \\
\metricpillorange{\small Focus}        & 0.61\textsuperscript{*}    & 7.48 $\pm$ 1.26 & 6.46 $\pm$ 2.09 \\
\metricpillorange{\small LitEngage}    & 0.61\textsuperscript{*}    & 7.62 $\pm$ 1.75 & 6.41 $\pm$ 1.72 \\
\metricpillorange{\small Refine}       & 0.60\textsuperscript{*}    & 7.31 $\pm$ 1.62 & 6.12 $\pm$ 2.32 \\
\bottomrule
\end{tabular}
\caption{Mixed effects modeling results from the expert user study. We report the standardized regression coefficient ($\beta$) and statistical significance (* $p<.05$, *** $p<.001$).
}
\label{tab:user_study}
\end{table}
\section{Conclusion}
We introduced \system, a framework for research idea evaluation that assesses ideas based on their soundness and contribution, grounded in scholarly literature. We also presented \benchmark, a multidisciplinary dataset of research ideas paired with expert-annotated review rubrics. Our experiments demonstrate that \system achieves higher coverage of points raised by human reviewers and consistently delivers higher-quality evaluations compared to baselines. Moreover, our user study shows a strong preference for \system as a useful idea evaluation tool, providing a confident indication of its potential to augment the ideation process in both AI and human workflows.

\section*{Limitations}
We recognize the following limitations and future work directions: 

\textbf{Limitations of \system.} First, as a literature-grounded framework, \system relies heavily on the retrieved literature to evaluate research ideas. Hence, it is possible that it might misjudge a method's effectiveness if it is not yet proven in the existing literature. \hnm{However, the vast majority of ideas are not entirely novel and have some precedents in the literature \citep{wang2017bias}.} 
Second, as we mention in Appendix \ref{app:results}, depending on the choice of the model backbone, running an idea evaluation using \system can take around 12 \textit{min} and cost up to \$3. \hnm{Although somewhat high, this remains on par with other literature search workflows \citep{skarlinski2024languageagentsachievesuperhuman, bragg2025astabenchrigorousbenchmarkingai}}. 
Future work may explore more lightweight versions of \system to enhance its usability to evaluate large batches of ideas. 

\textbf{Limitations of \benchmark and automatic evaluation.} We have made a considerable effort to select papers with high-quality reviews. However, idea evaluation remains a subjective task and human reviews cannot always be considered as ground-truth. Although the degree of coverage of human reviews can be one signal of evaluation comprehensiveness, it is not a definitive judge of overall quality. To that end, we have included other metrics in our setup to give a more holistic assessment, but the evaluation of open-form responses remains challenging, and there are other facets of evaluation that we have considered but that were challenging to scale or automate (e.g. citation factuality and relevance, usefulness, etc.). Additionally, our evaluation is limited to the four disciplines included in \benchmark, and although our framework is discipline-agnostic, our results might not necessarily generalize to other disciplines. Upon acceptance, we will release our entire dataset and evaluation pipeline, including the UI we used for the user study, seeking feedback from the wider scientific community on \system's utility.

\textbf{Limitations of Expert User Study.}
Some measures, such as those related to \metricpillorange{\small Focus}, are inherently subjective to the user, and could be influenced by stylistic factors rather than actual improved quality. Additionally, resarch idea quality is a source of variance, as participants were only allowed to submit unique research ideas to a given system. However, we collect a self-report of research idea detail, which is not statistically significantly different according to a Welch’s t-test. Additionally, the overall scores of our results may be inflated by our expert sample, who report a 7.2/10 on AI use for research. Our results may be different in a sample who use AI less often. This does not affect our relative improvement over o4-mini-deep-research, as we show concrete evidence that the study was adequately blinded (see \autoref{app:userstudy}). Future work should consider the barriers to adoption for automated research idea evaluation, especially for those who are less keen on using AI for their research cycle. Another strong direction would study how use of \system impacts short and long-term research success.
 
\section*{Ethical Considerations}
\benchmark is collected from papers on Openreview and eLife licensed under a Creative Commons Attribution (CC BY) license, which allows reuse with attribution. We provide the full list of papers used to create \benchmark in \autoref{app:benchmark}. 

\section*{Ethics Statement}
\benchmark is collected from papers on Openreview and eLife licensed under a Creative Commons Attribution (CC BY) license, which allows reuse with attribution. We provide the full list of papers used to create \benchmark in \autoref{app:benchmark}.

\section*{Acknowledgments}
This material is based upon work supported by the Ai2 Faculty Research Award. 
We thank Aaron Jencks, Jiwoo Park, and Yusen Peng for helping with data annotation of the quality metrics of \system and o4-mini-deep-research. We are also grateful to colleagues from Ai2 for their valuable feedback on an early version of the project, as well as colleagues from OSU NLP for their comments on the first draft of the manuscript. 

\section*{Author Contributions}

\begin{itemize}[left=0pt, labelsep=0.5em, itemsep=0.3em]
\item \textbf{Project leadership:} Hanane Nour Moussa, Patrick Queiroz Da Silva
\item \textbf{Project conception:} Bodhisattwa Prasad Majumder, Sachin Kumar, Hanane Nour Moussa, Patrick Queiroz Da Silva
\item \textbf{Development of \system:} Hanane Nour Moussa, Patrick Queiroz Da Silva
\item \textbf{\benchmark paper curation and LLM-based extraction:} Hanane Nour Moussa
\item \textbf{\benchmark expert validation:} Hanane Nour Moussa, Daniel Adu-Ampratwum, Alyson East, Zitong Lu, Nikki Puccetti, Mingyi Xue
\item \textbf{Evaluation pipeline and ablation experiments:} Hanane Nour Moussa
\item \textbf{Expert recruitment for user study:} Patrick Queiroz Da Silva, Hanane Nour Moussa
\item \textbf{\system interface for user study:} Patrick Queiroz Da Silva, Hanane Nour Moussa
\item \textbf{User study results analysis:} Patrick Queiroz Da Silva
\item \textbf{Manuscript writing and revision:} Hanane Nour Moussa, Patrick Queiroz Da Silva, Sachin Kumar, Huan Sun, Bodhisattwa Prasad Majumder
\item \textbf{Advisory:} Sachin Kumar, Bodhisattwa Prasad Majumder, Huan Sun
\end{itemize}



\bibliography{custom}

\appendix
\appendix
\section*{Appendix}


\addcontentsline{toc}{section}{Appendix}

We include herein details omitted from the main text as follows:
\begin{itemize}
  \item \hyperref[app:reproduce]{Reproducibility statement}
  \item \hyperref[app:system]{\system{} details and prompts}
  \item \hyperref[app:ui]{\system{} interface}
  \item \hyperref[app:benchmark]{\benchmark{} curation details}
  \item \hyperref[app:metrics]{Evaluation metrics details}
  \item \hyperref[app:exp]{Experimental setup details}
  \item \hyperref[app:results]{Evaluation results}
  \item \hyperref[app:userstudy]{User study details}
  \item \hyperref[app:examples]{\system{} output examples}
\end{itemize}

\section{Reproducibility statement}
\label{app:reproduce}
To ensure the reproducibility of our work, we provide full prompts and implementation details of \system (\autoref{app:system}), the detailed process of constructing \benchmark (\autoref{app:benchmark}), full information about our evaluation (\autoref{app:metrics}), experimental setup (\autoref{app:exp}), and expert user study (\autoref{app:userstudy}). Upon acceptance, we will also openly release our code and dataset.
\section{\system details and prompts}\label{app:system}
In this section we elaborate on implementation details omitted from the main text due to space limitations and present our full system prompts. 
\subsection{Soundness Details}
\subsubsection{Additional Implementation Details}
\textbf{Snippet Search.} We employ the \href{https://api.semanticscholar.org/api-docs/\#tag/Snippet-Text}{snippet search endpoint} from Semantic Scholar to get paper snippets (from the title, abstract, or body) that are relevant to the method being evaluated. To retrieve all papers referenced in the snippet, we use the field \texttt{refMentions} from the returned snippet data, which allows us to match every referenced paper to its Semantic Scholar \texttt{corpusID}. 

\textbf{Paper Downloading.} Once all referenced papers are identified, we use their \texttt{corpusID} to retrive their Semantic Scholar entry and use the field \texttt{OpenAccessPDF} to get the url to download the full text. Since there are cases where the paper might be behind a paywall, we use \href{https://unpaywall.org/}{Unpaywall} with our institution emails to retrieve open access version of the papers, if available. 

\textbf{Methods and Results Extraction.} To extract key sections from each paper, we first use GROBID to parse the paper PDF into its XML representation. Then, we search for the methods and results section by matching the section names to an exhaustive list of section titles that could potentially be used to reference the methods and results sections (e.g. Methods, Methodology, Protocol etc.). 

\textbf{TL;DR Summary.} Since the method-level soundness review can be lengthy, we also include a summarization step where we prompt an LLM to generate a TL;DR summary highlighting the top three most important strengths, weaknesses, and suggestions to address. In our \system user interface (Appendix \ref{app:ui}), both this summary and the method-level evaluation are shown to the user, with the latter being expandable. 

\textbf{Citation Checking.} As a post-processing step, we call a citation checking module that uses an LLM to perform the following functionalities: ensures that all citation worthy statements are followed by relevant citations and formats the bibliography section of the evaluation report. 
\subsubsection{Prompts}
In this section we provide all the prompts used in the Soundness Module.
\begin{tcolorbox}[enhanced,breakable,title=Methods Extraction, label=box:methods-extraction]

You are an expert research assistant. You are skilled at reading research ideas and identifying the methods that are being proposed to solve the research problem. Methods can be planned system designs, experiments, human studies, analyses, ablations, etc. 

\medskip

Given a research idea, you should extract all methods as a Python list, such that each method is a separate item in the list. Each item should be a word-for-word copy of a method, along with a short synthesis that grounds the method in the context of the overall research idea. The extracted methods should be interpretable on their own.

\medskip

Ensure that the methods you extract address different aspects of the research idea and are non-redundant.

\medskip

The method list you return should be ranked by importance to the research idea, with the most important methods first.

\medskip

[start research idea]  
\{research\_idea\}  
[end research idea]

\medskip

Please output a parseable Python block as follows:

\begin{lstlisting}[language=json]
```python
plans = ["context + method", ...]
```
\end{lstlisting}
\end{tcolorbox} 

\begin{tcolorbox}[enhanced,breakable,title=Query Generation, label=box:query-generation]

You are an expert research assistant. Given a method (i.e., one approach that researchers are adopting to execute their idea) extracted from a research idea, please construct a singular query that will be used to search for paper snippets using the Semantic Scholar API. 

\medskip

Use JSON format with 70 words or less per query. Do not include any text in the query besides the query itself. Do not include text like "semantic search query about ..." or "papers related to ...". Just the actual query text. No operators such as AND, OR should be used. Just a query in natural language that is relevant to the method.

\medskip

[start extracted method]  
\{clean\_method\}  
[end extracted method]

\medskip

In case the method does not have enough context to construct an effective snippet search query, you can use the research idea to understand the overall research direction and inject useful context.

\medskip

[start research idea]  
\{research\_idea\}  
[end research idea]

\medskip

Please output a parseable JSON block as follows, being especially careful to use the correct number of escape characters:

\begin{lstlisting}[language=json]
```json
{
    "query": "Your search query 
    here (IN 70 WORDS OR LESS)"
}
```
\end{lstlisting}
\end{tcolorbox}

\begin{tcolorbox}[enhanced,breakable,title=Method and Result Synthesis, label=box:method-result-synthesis]

You are an expert research assistant knowledgeable in many domains. You are extremely critical and observant, and do not overgeneralize findings. You are given a proposed research method and the methods/results section from a paper.

\medskip

[start paper]  
\{paper\_text\}  
[end paper]

\medskip

[start proposed research method]  
\{proposed\_method\}  
[end proposed research method]

\medskip

To further understand the scope of the proposed research method, here is the entire research idea that it is extracted from — a method is a single approach that researchers are adopting to execute their research idea:

\medskip

[start research idea]  
\{research\_idea\}  
[end research idea]

\medskip

For any method in the paper that is related to the proposed research method and the overall research idea, please summarize the method used in the paper, report the experimental outcome from using the method, and provide some context for experimental conditions.

\medskip

Do not use any in-text citations. Ensure that the method, results, and context you provide are specific and detailed, and that they mention how it relates to the proposed method and research idea.

\medskip

If the proposed research method does not relate to any methods in the paper, please return an empty dictionary.

\medskip

Strictly follow the output format displayed below.

\medskip

JSON formatting requirements:  
- Must be a complete, valid JSON object  
- Start with an open bracket and end with closed bracket  
- No trailing commas after the last property  
- Validate JSON structure before output  

\begin{lstlisting}[language=json]
```json
{
    "method": "Description of experimental approach including:
    algorithm/technique, datasets/inputs, computational resources, experimentation details, and evaluation setup, and metrics/instruments used, etc.",
    "results": "Quantitative outcomes with specific values,
    comparisons to baselines, statistical significance where
    applicable",
    "context": "Key experimental conditions:
    dataset/population size, hardware/system/instrument
    specs, hyperparameters, or other domain-specific
    constraints that affect reproducibility"
}
```
\end{lstlisting}
\end{tcolorbox}


\begin{tcolorbox}[enhanced,breakable,title=Method-Level Soundness Review Synthesis, label=box:method-soundness-review]

You are an expert research assistant knowledgeable in many domains. You are extremely critical and observant, and do not overgeneralize findings.
            
\medskip

You are given a proposed research method and a list of related work.

\medskip

Your objective is to create a meta-review of the related work in the context of the proposed research method. Point out any evidence that supports or contradicts the proposed method. Make sure to contrast the related work as a series of iterative scientific work, where newer work can provide evidence that supports or contradicts older work.

\medskip

It is important that the meta-review you generate always ties back to the original research idea. Judge the support and contradictions as well as suggested actions for each method within the general context of the research idea to ensure that your review is highly relevant and precise.

\medskip

Be granular, making sure to reference specific details such as:
\begin{itemize}
  \item algorithm/technique, datasets/inputs/population, computational resources, statistical methods, implementation details, and evaluation setup, and metrics/instruments used, etc.
  \item quantitative outcomes, comparisons to baselines, statistical significance
  \item dataset/population size, hardware/system specs, hyperparameters, or other domain-specific constraints that affect reproducibility
\end{itemize}

It is important that for each method-level meta-review that you generate, your review of the support and contradictions should be ordered starting from strongest evidence of support/contradiction to the weakest. Likewise, the suggested actions should be ordered from most important to least important. This does not mean that you will generate these as bullet points, but rather detailed, coherent paragraphs that are logically ordered.

\medskip

It is required to copy the in-text citations with their links in Markdown format \texttt{[(author, YYYY-MM)](link)} when referring to related work.

\medskip

\textbf{Related Work:}  
\texttt{\{related\_work\}}

\medskip

[start proposed research method]  
\texttt{\{pm\}}  
[end proposed research method]

\medskip

[start research idea]  
\texttt{\{rp\}}        
[end research idea]

\medskip

\textbf{Output format:}  
Please output a parseable JSON block as follows:

\begin{lstlisting}[language=json]
```json
{
    "support": "evidence that supports the proposed method",
    "contradictions": "evidence that contradicts the proposed 
                       method",
    "suggested_action": "how can the proposed method be improved
                        based on the related work",
    "soundness_score": "int score 0 to 10 based on the evidence 
                        for and against the proposed method"
}
```
\end{lstlisting}

\end{tcolorbox}

\subsection{Contribution Details}
\subsubsection{Additional Implementation Details}
\textbf{Paper Search.} To retrieve relevant papers for contribution analysis, we use the \href{https://api.semanticscholar.org/api-docs/#tag/Paper-Data/operation/get_graph_paper_relevance_search}{Semantic Scholar paper relevance search} which returns the top n most relevant papers based on a query. For subsequent processing in the contribution module (i.e. relevance assessment and pairwise comparison) we retrieve the paper abstract from the returned \texttt{abstract} field. 

\textbf{Paper Downsampling.} Before the pairwise comparison step, we sample 25 papers from the final list. This downsampling serves two functions: first it reduces the latency and computational cost of the pairwise comparison step, and it reduces the overall context forwarded to the final synthesis step to avoid prohibitive context lengths. 

\textbf{Citation Checking.} Similar to the Soundness module, we also apply a final post-check on the citations to ensure proper attribution and bibliography formatting. 
\subsubsection{Prompts}
In this section we provide all the prompts used on the Contribution Module. 
\begin{tcolorbox}[enhanced,breakable,title=Dimension Extraction Prompt, label=box:dimension-extraction]

You are a helpful assistant that reads scientific research ideas and extracts a structured summary of their contributions. 

\medskip

Your task is to identify a small number of high-level contribution dimensions, and for each dimension, extract one or more specific contribution statements that are faithful to the research idea.

\medskip

Contribution dimensions should represent general categories of scientific contribution that are meaningful and comparable across research ideas, regardless of the field. These might include:  
- methodology (e.g., proposing a new method, model, or procedure)  
- application (e.g., applying existing methods to a new problem or domain)  
- theoretical contribution (e.g., proving a new result, deriving a new model)  
- data (e.g., constructing a new dataset, conducting original measurements or surveys)  
- evaluation (e.g., designing an experimental protocol, benchmarking a technique)  
- tool or system design (e.g., building software, devices, or infrastructure to support research)  
- conceptual framework (e.g., introducing a new taxonomy or way of thinking about a problem)  

\medskip

Do not limit your output to the examples above but rather generate suitable dimensions for the research idea given to you. Only include dimensions that are actually reflected in the research idea — do not add generic or speculative categories. 

\medskip

For each dimension, write one or more contribution statements that clearly explain what the research is proposing. These statements should be precise, self-contained, and informative — make sure they include enough context as they will be used as the basis to generate search queries later on.

\medskip

Each dimension may include multiple contribution statements, but do not repeat the same idea across dimensions. Avoid redundancy, and keep the summary compact and informative. 

\medskip

Please output a parseable JSON block as follows:

\begin{lstlisting}[language=json]
```json
{
  "<dimension_name_1>": [
    "<contribution_statement_1>",
    "<contribution_statement_2>"
  ],
  "<dimension_name_2>": [
    "<contribution_statement_3>"
  ]
}
```
\end{lstlisting}

\end{tcolorbox}

\begin{tcolorbox}[enhanced,breakable,title=Query Generation (Contribution), label=box:query-generation-2]

You are an expert at writing highly targeted search queries for retrieving academic papers using the Semantic Scholar API. 

\medskip

You will be given a full research idea and one specific contribution from that idea. 

\medskip

Your task is to generate up to \{n\_queries\} short, diverse, and high-quality search queries that are focused on the given contribution and consistent with the overall research context.

\medskip

These queries will be used to search for paper abstracts. They must be semantically rich and optimized to retrieve papers that are directly relevant to the contribution. 

\medskip

Avoid generating queries that are too general, overly broad, or likely to return unrelated results. Also avoid generating queries that are just broadly related to the research idea but not specifically tailored to the contribution.

\medskip

\textbf{Guidelines:}
\begin{itemize}
    \item Each query should be brief and focused (as if typed into a search bar; as a rule of thumb do not exceed 7 words per query).
    \item Queries must stay tightly aligned with the core idea of the contribution.
    \item Incorporate key methods, problems, domains, or goals described in the contribution.
    \item Reflect an understanding of the broader research idea, but do not drift away from the specific contribution.
    \item Use natural phrasing (no Boolean operators like AND, OR, etc.).
    \item Ensure queries are meaningfully different from one another while remaining on-topic.
\end{itemize}

\end{tcolorbox}

\begin{tcolorbox}[enhanced,breakable,title=Relevance Assessment Prompt, label=box:relevance-assessment-1]

You are an expert at evaluating whether a paper should be considered relevant for assessing the scientific contribution of a research idea. 

You will be given a research idea and a paper abstract retrieved from Semantic Scholar. Your task is to thoroughly understand the scientific contributions of each of the research idea and the paper, and to output a score from 0 to 5 indicating how similar the contributions of the paper are to those of the research idea.

\medskip

\textbf{Scoring Rubric:}

\textbf{Score 5 — Highly Similar Contributions:}
\begin{itemize}
    \item The paper addresses the exact same research question or hypothesis as the idea
    \item Uses identical or very similar methodological approaches
    \item Targets the same specific population, system, or domain
    \item Would directly compete with or overlap significantly with the proposed research
    \item The paper's findings would substantially impact the novelty of the proposed work
\end{itemize}

\textbf{Score 4 — Very Similar Contributions:}
\begin{itemize}
    \item The paper addresses a closely related research question with significant overlap
    \item Uses similar methodological approaches with minor variations
    \item Targets a very similar population, system, or domain
    \item Shares most key variables, measurements, or outcomes of interest
    \item The paper's contributions would moderately impact the proposed research's novelty
\end{itemize}

\textbf{Score 3 — Moderately Similar Contributions:}
\begin{itemize}
    \item The paper addresses a related research question within the same broad area
    \item Uses some similar methods or approaches but with notable differences
    \item Targets a related but distinct population, system, or domain
    \item Shares some key concepts, variables, or theoretical frameworks
    \item The paper provides useful context but doesn't directly threaten novelty
\end{itemize}

\textbf{Score 2 — Somewhat Similar Contributions:}
\begin{itemize}
    \item The paper is in the same general field or discipline
    \item Uses different methods but addresses conceptually related problems
    \item Limited overlap in specific research focus or target populations
    \item Shares broad theoretical background but differs in specific contributions
    \item The paper is peripherally relevant for background or context
\end{itemize}

\textbf{Score 1 — Minimally Similar Contributions:}
\begin{itemize}
    \item The paper is tangentially related to the research area
    \item Very limited overlap in methods, populations, or specific research questions
    \item May share some terminology or broad field classification
    \item Provides minimal insight relevant to the proposed research
    \item Connection is primarily at the disciplinary level
\end{itemize}

\textbf{Score 0 — No Similar Contributions:}
\begin{itemize}
    \item The paper addresses completely different research questions
    \item No meaningful overlap in methods, populations, or domains
    \item Different field or discipline entirely
    \item No relevant insights for the proposed research
    \item No discernible connection between the contributions
\end{itemize}



Respond with a JSON object containing two fields \texttt{rationale} and \texttt{score}.  
Please output a parseable JSON block as follows:

\begin{lstlisting}[language=json]
```json
{
  "rationale": "<your detailed reasoning for the score>",
  "score": <an integer from 0 to 5 based on the rubric>
}
```
\end{lstlisting}

\end{tcolorbox}
\begin{tcolorbox}[enhanced,breakable,title=Pairwise Comparison Prompt, label=box:pairwise-comparison]

You are an expert in assessing the novelty of scientific research ideas relative to existing work.

Your job is to compare a full research idea to a paper abstract based on novelty and originality.

You will be given a research idea, a paper abstract, and a comma-separated list of contribution dimensions.

\medskip

Produce a structured output consisting of:

\begin{enumerate}
  \item \texttt{overall\_comparison}: a broad yet precise summary of the novelty of the research idea versus the paper, i.e., whether the idea proposes ideas/angles/uses that appear original relative to what the abstract claims; identify overlap vs. originality explicitly.
  \item \texttt{dimension\_comparisons}: for each provided dimension, a novelty comparison that states whether the idea is doing something not present in the paper for that dimension (or vice versa), and a numeric score:
  \begin{itemize}
    \item 1 = The research idea is more novel under this dimension (adds ideas/angles/uses not present in the paper abstract)
    \item 0 = Neither appears more novel or the paper does not address this dimension (tie/insufficient evidence)
    \item -1 = The paper appears more novel or the idea largely replicates what the paper already presents under this dimension
  \end{itemize}
\end{enumerate}

\medskip

Return your output as a parseable JSON block with exactly this structure:

\begin{lstlisting}[language=json]
```json
{
  "overall_comparison": "<novelty-focused summary>",
  "dimension_comparisons": {
    "<dimension_1>": {
      "comparison": "<comparison for this dimension>",
      "score": <1 | 0 | -1>
    },
    "<dimension_2>": {
      "comparison": "<comparison>",
      "score": <1 | 0 | -1>
    }
  }
}
```
\end{lstlisting}

\end{tcolorbox}

\begin{tcolorbox}[title=Dimension-level Contribution Synthesis Prompt, label=box:dimension-synthesis]

You are reviewing a research idea for novelty and originality of its contributions and their impact. 

You are critical, knowledgeable, precise, and not afraid to give truthful assessments of the idea's novelty. Do not discuss methodology soundness, feasibility, correctness, or experimental quality — those are out of scope here.

\medskip

You will be given a full research idea and a JSON file of pairwise comparisons between the idea and existing papers. Each comparison includes a paper reference (authors, title, year, URL), an \texttt{overall\_comparison} (novelty-focused), and \texttt{dimension\_comparisons} comparing the research idea to a related work based on a specific contribution dimension.

\medskip

Your task is to synthesize an overall evaluation of the idea's novelty and impact based on all pairwise comparisons.

\medskip

\textbf{Structure:}

Produce a novelty and impact assessment for each contribution dimension found in the comparisons. Each paragraph focuses on one dimension and contains three subsections:

\begin{itemize}
  \item \textbf{Strengths:} Precisely explain where the idea is novel under this dimension versus the papers and makes an impact; back claims with in-text citations.
  \item \textbf{Weaknesses:} Precisely explain where the idea lacks novelty or is already covered by prior work, and hence lacks impact; back claims with in-text citations. Be exhaustive with identifying weaknesses.
  \item \textbf{Suggestions:} Give actionable, feasible, and useful suggestions to improve the novelty and impact of the work, if needed, based on the evidence from the Strengths and Weaknesses sections; back claims with in-text citations.
\end{itemize}

Use in-text citations with links in Markdown format: \texttt{[Author et al., Year](URL)}. 

\begin{lstlisting}[language=json]

\end{lstlisting}

\end{tcolorbox}

\section{\system interface}\label{app:ui}
\begin{figure*}[ht]  
    \centering
    \includegraphics[width=1\linewidth]{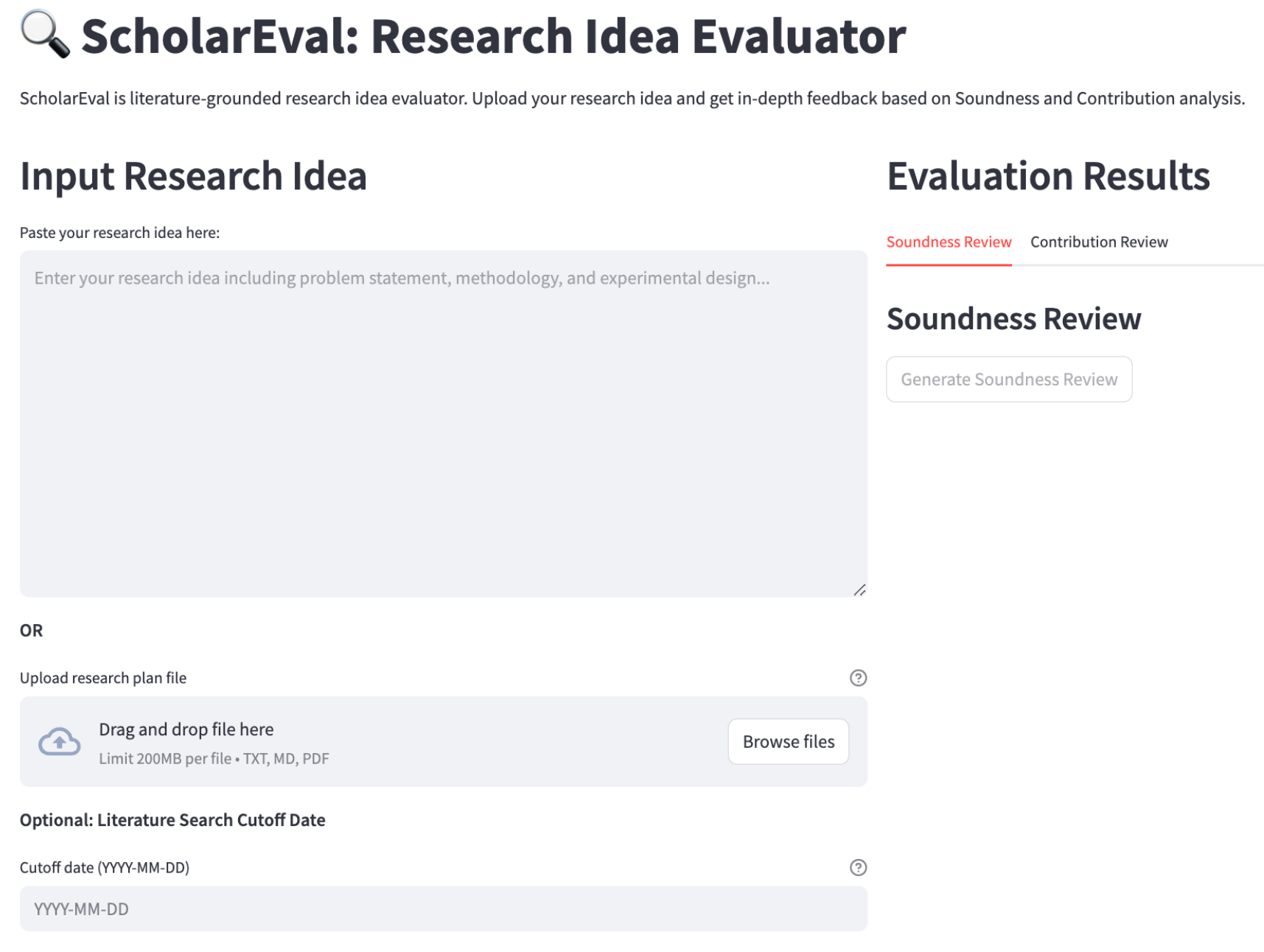}
    \caption{\system user interface.}
    \label{fig:ui}
\end{figure*}

To conduct the user study and for wider feedback from the community (upon acceptance), we create a \system user interface. As shown in \autoref{fig:ui}, our interface includes an input box for the user to enter their research idea or upload it from a file. Optionally, the user can specify a literature search cutoff date, This is especially useful if \system is used to evaluate an already published research idea. The user can then generate a Soundness review or switch tabs to generate a Contribution review. 

\section{\benchmark curation details}\label{app:benchmark}
\subsection{Paper List}
We provide the full list of the 117 papers that were used to construct \benchmark at the end of the manuscript. 
Each entry in the lists links to the paper on OpenReview or eLife.
\subsection{LLM-based extraction}
\subsubsection{Details}
For each paper in our dataset, 
we use Claude 4 Sonnet to extract the research idea and review rubrics. The extraction of the review rubrics is done in two stages: first, we instruct the LLM to extract the verbatim excerpts from the review that give assessments about the research idea. In the second step, we instruct the LLM to remove redundancies and format the extracted text into standalone statements classified based on type, axis, and severity. The full prompts used in this process are provided in section \ref{app:extraction-prompts}. 

\subsubsection{Prompts}\label{app:extraction-prompts}

\begin{tcolorbox}[enhanced,breakable,title=Research Idea Extraction, label=box:research-idea-extraction]

You are a highly skilled assistant tasked with thoroughly reading the content of a research paper and extracting its research idea. The research idea should reflect the state of the research at the time it was proposed, before any experiments or conclusions were drawn.

\medskip

A research idea consists of the following key sections:

\textbf{Problem:}
\begin{itemize}
  \item This section provides the background of the problem being addressed, the motivation for pursuing it, and any hypotheses that the authors have formed at the start of their research.
  \item Focus on the initial framing of the problem, why it's important, and the research questions that guide the study.
\end{itemize}

\textbf{Method:}
\begin{itemize}
  \item Summarize the methodology that the authors will follow to address the problem. This includes the overall approach, any theoretical frameworks, models, or algorithms to be used, and the rationale behind the chosen methodology.
  \item Do not include any references to specific results or findings — only the proposed methods and strategies.
\end{itemize}

\textbf{Experiment Design:}
\begin{itemize}
  \item Describe the experiments that the authors will conduct to test their hypotheses. This should include the design, variables, procedures, and any tools or techniques that will be employed.
  \item Provide details about how the authors plan to gather data, measure outcomes, and assess the effectiveness of the methods, but do not include any results.
  \item However, make sure not to make any reference to experimental setups that the authors could not have known before conducting the experiments (e.g., specific values for hyperparameters or other design choices that were made through trial and error).
\end{itemize}

\textbf{Key Constraints:}
\begin{itemize}
  \item \textbf{No results or conclusions:} Exclude any findings, outcomes, or conclusions that were reached after conducting the experiments. The research idea should represent the research at the proposal stage, not the completion stage.
  \item \textbf{First-person perspective:} Write the research idea from the authors' point of view, as if the authors themselves are outlining their proposed work. Avoid referring to the authors in the third person.
  \item \textbf{Concise and focused:} The research idea should be clear, concise, and direct. Include only the information necessary to describe the research as it was proposed, not the outcomes.
  \item \textbf{Represent the status before experimentation:} Ensure the idea reflects the status of the project when it was still in the idea or proposal phase, prior to conducting any experiments.
\end{itemize}

\textbf{General Writing Guidelines:}
\begin{itemize}
  \item Use active voice as if it's written by the authors themselves.
  \item Maintain a formal yet straightforward tone typical of research ideas.
  \item Keep the text brief and to the point, and avoid extraneous details or unnecessary elaboration.
\end{itemize}

The research idea should be faithful to the original research paper, concise, and aligned with the intent of the authors at the time of the project ideation phase.

\medskip

Generate the research idea for the following paper:

\begin{lstlisting}[language=json]
{paper_content}   
\end{lstlisting}

\end{tcolorbox}

\begin{tcolorbox}[enhanced,breakable,title=Verbatim Review Extraction, label=box:verbatim-review-extraction]

You are an expert assistant skilled at extracting key components of scientific paper reviews containing evaluations of the underlying research idea and discarding all other criticism of other aspects of the scientific paper, including results and presentation. 

Namely, you will be given a text containing all the reviews that the scientific paper received as well as a research idea detailing the main research idea contained in the scientific paper, and your job is to extract sentences from the review that address components mentioned in the research idea. Any sentences that contain criticism of aspects of the paper not mentioned in the research idea should not be extracted. This is an important guideline to follow. 

\medskip

\textbf{Reviews contain the following components:}
\begin{itemize}
    \item \textbf{Summary:} A brief overview of the paper's contributions, main ideas, and objectives.
    \item \textbf{Strengths:} Positive aspects highlighted by the reviewer, such as innovative approaches, solid methodology, or valuable contributions.
    \item \textbf{Weaknesses:} Negative aspects or concerns raised by the reviewer, such as lack of clarity, methodological flaws, or unanswered questions.
    \item \textbf{Questions:} Queries or points of clarification raised by the reviewer, often regarding experimental design, assumptions, or aspects of the research idea.
\end{itemize}

\textbf{The research idea corresponding to the research paper includes:}
\begin{itemize}
    \item \textbf{Problem:} Background information on the problem the paper addresses, the motivation for pursuing it, and any hypotheses formed by the authors.
    \item \textbf{Method:} The methodology proposed by the authors to address the problem.
    \item \textbf{Experiment Design:} Details on the experiments, data collection, and evaluation techniques that the authors plan to employ.
\end{itemize}

\medskip

\textbf{Review:}
\begin{lstlisting}[language=json]
{review_text} 
\end{lstlisting}

\textbf{Research Idea:}
\begin{lstlisting}[language=json]
{research_idea_text}    
\end{lstlisting}

\medskip

\textbf{Output format:}  
This is an extraction task, not a generation task. Your job is exactly to extract sentences from the reviews which address the ideation behind the research (i.e., the research at the idea stage, imagine that no results have been obtained or paper written yet).  

These should only include criticism related to the problem that the scientists are trying to solve, the methodology they are employing, or the experiments they plan to conduct. Any other content related to typographical errors, presentation or formatting issues, results, comments on figures and tables, artifacts, or conclusions derived from executing the experiments should absolutely not be included.  

Only points related to the underlying research idea as specified in the research idea should be kept. Any point not related to an aspect explicitly mentioned in the research idea should not be included.  

Return the extracted sentences (\textbf{verbatim! no changes made}) as follows: 

\begin{lstlisting}[language=json]
Extracted Review: <verbatim extracted sentences>    
\end{lstlisting}

Nothing else besides the extracted review should be contained in the response.

\end{tcolorbox}

\begin{tcolorbox}[enhanced,breakable,title=Review Formatting Into Rubrics, label=box:review-rubrics-1]

You are an expert assistant skilled at extracting clear and non-redundant evaluation statements from a review summary. In this task you will be given a summary that was created by merely concatenating extracted sentences from a longer review. You will be given this extractive review summary along with the research idea that it reviews and you have to do the following tasks while taking into account the constraints.

\medskip

\textbf{Tasks:}
\begin{itemize}
  \item \textbf{Removing redundancies:} The review summary was created by concatenating the extracted sentences from multiple reviews written by different reviewers about the same paper. Thus, some of the sentences can be redundant. You should remove all redundancies. If a point is mentioned more than once, drop the other mentions and only keep one.
  \item \textbf{Express each evaluation} (weakness or strength mentioned by the reviewer) as a clear statement. Hence your final output should be a list of such statements, each one addressing a specific point made by the reviewer(s).
  \item \textbf{Adding context:} The review summary might refer to some aspects of the research idea without sufficient context. You should aim to make each statement in the list clearer by adding context from the research idea that would enhance the understanding of the review.
  \item \textbf{Remove mentions of the final paper:} These statements are extracted from the reviews of a paper submitted to a conference or journal. Hence, some statements might have mentions of the final paper (e.g., “in line so and so …” or “in Figure x …”). When such statements are encountered, you should slightly rephrase them — without alterations to the meaning — to ensure that they do not refer to the paper but rather to the corresponding component in the research idea. Ensuring this consistency between the statements and the research idea content is crucial.
  \item For each statement (item in the list), you should also annotate it with whether it is a \texttt{"strength"} or a \texttt{"weakness"}, whether it is related to the paper’s \texttt{"soundness"} (correctness of the methods, etc.) or \texttt{"contribution"} (novelty and contribution to the field), and finally annotate whether it represents a \texttt{"major"} or \texttt{"minor"} point. Major points are those that reviewers highly stress and that affect the overall evaluation of the paper, while minor points are those that are mentioned but not stressed as much.
\end{itemize}

\textbf{Constraints:}
\begin{itemize}
  \item Make sure that the changes you make are minimal and only aim to remove redundancy, express the statements as a list, add context, and ensure consistency with the research idea content.
  \item Do not completely paraphrase the review summary. Try to keep the original wording and sentences as much as possible. Your changes should not lead to a significant overhaul of the review summary.
  \item Do not change the tone of the review. It should still be written as if produced by an actual reviewer.
\end{itemize}

\medskip

\textbf{Extractive review summary:}
\begin{lstlisting}[language=json]
{intermediate_text}   
\end{lstlisting}

\textbf{Research Idea:}
\begin{lstlisting}[language=json]
{research_idea_text}    
\end{lstlisting}



Please output a parseable JSONL (JSON Lines) block where each item in the JSONL is a JSON object on a single line as follows:

\begin{lstlisting}[language=json]
```jsonl
{
  "statement": "<one clear, self-contained statement>", 
  "type": "strength",
  "axis": "soundness", 
  "severity": "major"
}
{
  "statement": "<another clear, self-contained statement>", 
  "type": "weakness", 
  "axis": "contribution", 
  "severity": "minor"
}
{
  "statement": "<another clear, self-contained statement>",
  "type": "strength", 
  "axis": "soundness", 
  "severity": "minor"
}
\end{lstlisting}
\end{tcolorbox}

\subsection{Expert Validation}
\subsubsection{Details}
We invited six experts of Ph.D. students, postdoctoral researchers, and professors (1 from computer science, 2 from biochemistry, 2 from neuroscience, and 1 from ecology). We first conducted a training session to explain the validation task and we have provided the annotators with detailed instructions, shown in section \ref{app:validation-instructions}. We made an effort to match every annotator to research ideas in their specific area of specialty, and annotators were instructed to skip a research idea if they felt that it fell out of their area of expertise. 

\subsubsection{Validation Instructions}\label{app:validation-instructions}

\begin{tcolorbox}[enhanced,breakable,title=\benchmark Annotation Guidelines, label=box:annotation-guidelines]

\textbf{Get familiar with the data}

Each task contains a research idea (.txt file) and a review summary (.jsonl file).

The research idea is composed of three main sections: 
\begin{itemize}
  \item \textbf{Problem:} the problem that the research idea tackles
  \item \textbf{Methodology:} the adopted methodology
  \item \textbf{Experiments:} the planned experimental setup
\end{itemize}

On the first line of the research idea, you will find a link to the original publication and public reviews on eLife. This is the link to the first version of the publication (v1), which contains the publication before any changes or revisions. On the eLife website, the publication is available under the \textit{Full text} tab and the full reviews under the \textit{Peer Review} tab.

The review summary is a JSONL file. Each line in the file is a dictionary structure with the following attributes:
\begin{itemize}
  \item \textbf{Statement:} a statement extracted from the original reviews
  \item \textbf{Type:} whether the statement represents a `weakness` or `strength`
  \item \textbf{Axis:} whether the statement is related to the `soundness` of the work (correctness and validity of the methods) or its `contribution` (novelty and impact)
  \item \textbf{Severity:} whether this is a `major` point (emphasized by the reviewer) or a `minor` one
\end{itemize}

\medskip
\textbf{Annotate a task}

For each task:  
\begin{enumerate}
  \item \textbf{Check the research idea:}  
  Ensure it represents the state of the work at the ideation stage, i.e. before any experiments were conducted or results obtained. It should only contain the problem, methodology, and planned experiments that the authors could have known at the ideation stage. Verify that the content is faithful to the original publication without significant alterations.  

  \item \textbf{Check the review summary:}  
  The statements should only pertain to the underlying research idea. Points about results, details only known after experimentation, figures, tables, or presentation issues should not be included.  
  Verify that:  
  \begin{itemize}
    \item The statements pertain only to the research idea (problem, methodology, planned experiments).  
    \item All such statements are extracted.  
    \item Each statement type is correctly categorized as `strength` or `weakness`.  
    \item Each statement axis is correctly categorized as `soundness` or `contribution`.  
    \item Each statement severity is correctly categorized as `major` or `minor`.  
  \end{itemize}

  \item \textbf{Make necessary changes:}  
  If the above guidelines are not respected, make modifications (e.g., editing the research idea, adding/removing statements in the review summary, or correcting type/axis/severity labels).
\end{enumerate}

\end{tcolorbox}

\subsubsection{Validation Results}
Table \ref{tab:expert-edits} shows the distribution of edits for each of the research idea and review rubrics. Common issues that required edits include the mention of results in the research idea, errors in the classification of statements based on their severity, and the inclusion of rubrics mentioning the results obtained in the paper or the referring to the presentation (tables, figures, etc.). 
\begin{table}[ht]
\centering
\small
\begin{tabular}{l c}
\toprule
\textbf{Category} & \textbf{Percentage of Edits} \\
\midrule
Research Idea   & 13.40\% \\
Review Rubrics  & 79.10\% \\
\bottomrule
\end{tabular}
\caption{Distribution of edits across research ideas and review rubrics.}
\label{tab:expert-edits}
\end{table}
\section{Evaluation metrics details}\label{app:metrics}
\subsection{Coverage Metric}
To compute the coverage metric, we use Prometheus-Eval \citep{kim2024prometheus} framework with GPT-4 \citep{openai2024gpt4technicalreport} backbone. We provide Prometheus with the detailed rubric given below to give a score on the degree of coverage of each rubric of \benchmark. These scores are subsequently averaged to compute the final Coverage score. 
\begin{tcolorbox}[enhanced,breakable,title=Coverage rubric given to Prometheus, label=box:coverage-rubric-prometheus]
\begin{lstlisting}[language=json]
rubric_data = { 
  "criteria": 
  "You are given a single essential statement from 
  a research idea review targeting the technical 
  soundness and scientific contribution of a research 
  idea. This statement was extracted from actual human 
  reviews about the research idea and will be used 
  as your ground-truth reference. Additionally, you 
  will be given a research idea review that aims to 
  give a detailed evaluation of the same research idea. 
  The review given to you will typically be long and 
  composed of multiple sections and references to related 
  papers. Your job is to verify that the given review 
  has a mention of the reference statement in its content. 
  These references can be paraphrased or articulated 
  differently, but they should express the same meaning 
  as the reference statement. Does the given review 
  express the reference statement?",

  "score1_description": 
  "The given review does not mention the reference 
  statement at all",

  "score2_description": 
  "The given review mentions some related concepts 
  but does not express the core meaning of the 
  reference statement",

  "score3_description": 
  "The given review partially expresses the reference 
  statement but misses key aspects or nuances",

  "score4_description": 
  "The given review expresses most aspects of the 
  reference statement with minor omissions or slight 
  differences in emphasis",

  "score5_description": 
  "The given review clearly expresses the reference 
  statement, capturing its core meaning and implications"
} 
\end{lstlisting}

\end{tcolorbox}
\subsection{Reference Invalidity}
We compute reference invalidity as the fraction of non-resolving citations by issuing an HTTP HEAD request for every paper link referenced in the output of \system and baselines, and inspecting the returned status code. In our initial implementation, we noticed that this approach undercounts failures: many publishers return 403 (bot blocking) or even 200 (soft-404 pages etc.) for URLs that do not actually resolve to the referenced papers upon manual inspection. We thus resort to reporting a lower bound for reference invalidity; we label a link as invalid only when it returns 404, 410, or any 5xx status. However the true of reference invalidity for baseline systems is higher. We provide failure cases observed in our manual audit in \Sref{app:manual-audit}. Reference invalidity is entirely eliminated in \system.
\subsection{LLM Metrics}
We use Claude 4 Sonnet as our LLM judge and instruct it to choose the winner between a pair of reports (\system and o4-mini-deep-research) using the prompt below. 

To validate the reliability of the LLM judgments, we conduct a small scale study with 33 report pairs and ask 6 Ph.D. students in artificial intelligence, neuroscience, and chemistry - who are unfamiliar with our system and would thus not be able to identify it in the pair - to choose the better report based on the three criteria. We provide our annotators with the same instructions given to the LLM judge. Each report pair was rated by 1, 2, or 3 annotators, and we set the final human label for each report pair based on majority vote. Reports that had equally split ratings were discarded, resulting in 18 remaining pairs. We compute the inter-annotator agreement between the human and LLM labels based on percent agreement and Gwet's AC1, which is a suitable inter-annotator agreement metric for cases representing class imbalance. The results shown in Table \ref{tab:iaa_results} underscore a high percent agreement for all metrics and a moderate to substantial agreement based on Gwet's AC1. 

Additionally, we provide the human preference results based on the full sample of 33 report pairs in Table \ref{tab:human-preference}. 
These preference results showcase that \system is substantially preferred over OpenAI Deep Research across all three metrics and corroborate the LLM-judge preference results. 

\begin{tcolorbox}[enhanced,breakable,title=LLM judge prompt, label=box:llm-judge]

You are an expert scientific reviewer. You will be given two research evaluation reports (Report A and Report B) and you must answer which one is better based on the following criteria:

\medskip

\textbf{Evidence support:}  
An evidence-supported report is one that grounds its claims in the literature and backs them with relevant citations that improve traceability and its overall trustworthiness and reliability. Which report has more evidence support?

\medskip

\textbf{Actionability:}  
Actionability is the extent to which the report offers varied, clear, and actionable suggestions that are likely to improve the research idea. Which report is more actionable?

\medskip

\textbf{Depth:}  
Depth is measured by the degree of engagement with the point being evaluated and the literature cited. A deep report discusses each point from multiple angles and references specific details about the literature it cites, rather than relying on generic statements followed by citations. Which response has greater depth?

\medskip

You must choose either \texttt{A}, \texttt{B}, or \texttt{Tie} (when both responses are equivalent). Read both responses thoroughly before judging, and give thorough rationales that show that you are a fair and unbiased judge. If you choose that one response is better, clearly rationalize why. If they are equivalent, then you should choose \texttt{Tie}.

\medskip

\textbf{Report A:} \texttt{\{prompt\_report\_a\}}  
\textbf{Report B:} \texttt{\{prompt\_report\_b\}}

\medskip

Judge the responses based on the criteria, while respecting the output format below:

\begin{lstlisting}[language=json]
```json
{
  "Evidence-support-rationale": "...", 
  "Evidence-support-winner": "<A, B, or Tie>", 
  "Actionability-rationale": "...", 
  "Actionability-winner": "<A, B, or Tie>", 
  "Depth-rationale": "...", 
  "Depth-winner": "<A, B, or Tie>"
}
\end{lstlisting}

\end{tcolorbox}
\begin{table*}[h]
\centering
\small
\begin{tabular}{l c c}
\toprule
\textbf{Metric} & \textbf{Percent Agreement (\%)} & \textbf{Gwet's AC1} \\
\midrule
Evidence support & 66.67 & 	0.4476 \\
Actionability    & 72.22 & 0.1667 \\
Depth            & 61.11 & 0.3883 \\
\bottomrule
\end{tabular}
\caption{Inter-annotator agreement (human vs.\ LLM) across metrics.}
\label{tab:iaa_results}
\end{table*}
\begin{table*}[h]
\centering
\small
\begin{tabular}{lccc}
\toprule
\textbf{Metric} & \textbf{\system} & \textbf{OpenAI Deep Research} & \textbf{Tie} \\
\midrule
Evidence Support & 70.6 & 14.7 & 14.7 \\
Depth            & 79.4 & 11.8 & 8.8  \\
Actionability    & 82.4 & 11.8 & 5.9  \\
\bottomrule
\end{tabular}
\caption{Human preference results on sample of 34 report pairs. }
\label{tab:human-preference}
\end{table*}

\section{Experimental setup details}\label{app:exp}
\subsection{Baselines}
In this section we provide the detailed prompts used to generate the Soundness and Contribution evaluation for all baseline systems. To ensure a fair comparison, we use optimized prompts that instruct all baseline systems to generate detailed results in the same format as \system. 

Similar to \citet{li2025reportbenchevaluatingdeepresearch}, to avoid cases where the retrieval-augmented models (i.e. GPT-4o-search-preview and o4-mini-deep-research) retrieve the paper corresponding to the research idea being evaluated, we include an instruction to limit the retrieved literature to publications released before the cutoff date (i.e. the publication date of the paper that the research idea was extracted from). Although not error proof, in our experimentation we have observed that this greatly reduces the instances where the target paper is retrieved.

\begin{tcolorbox}[enhanced,breakable,title=Baselines Soundness Evaluation Prompt, label=box:baselines-soundness]

You are an expert research assistant. You are skilled at evaluating the soundness of all methods proposed in a research idea.

\medskip

Create an evaluation for every method being proposed to solve the research problem. If there are \texttt{n} methods there should be \texttt{n} evaluations. Methods can be planned system designs, experiments, human studies, analyses, ablations, etc.
            
\medskip

You should find related work for each method being proposed, such that it can provide evidence that supports or contradicts the method at hand.

\medskip

It is important that the evaluation you generate always ties back to the original research idea. Judge the support and contradictions as well as suggested actions for each method within the general context of the research idea.

\medskip

Be granular, making sure to reference specific details such as:
\begin{itemize}
  \item algorithm/technique, datasets/inputs, computational resources, implementation details, and evaluation setup, and metrics used
  \item quantitative outcomes, comparisons to baselines/state-of-the-art, statistical significance
  \item dataset size, hardware specs, hyperparameters, or other domain-specific constraints that affect reproducibility
\end{itemize}

It is required to put the in-text citations with their links in Markdown format \texttt{[(author, YYYY-MM)](link)} when referring to related work.

\medskip

Here is the research idea to evaluate:

[start research idea]  
\texttt{\{research\_idea\}}  
[end research idea]

\medskip

\textbf{The output should be formatted as follows, such that each extracted method from the research idea would be its own section title, in which there are three subsections: strengths, weaknesses, and suggestions.}

\begin{lstlisting}[language=json]
<method_1>
"support": "evidence that supports the proposed method",
"contradictions": "evidence that contradicts the proposed 
method",
"suggestions": "how can the proposed method be improved based on 
the related work"

<method_2>
"support": "evidence that supports the proposed method",
"contradictions": "evidence that contradicts the proposed 
method",
"suggestions": "how can the proposed method be improved based on 
the related work"

<method_n>
"support": "evidence that supports the proposed method",
"contradictions": "evidence that contradicts the proposed 
method",
"suggestions": "how can the proposed method be improved based 
on the related work"
\end{lstlisting}



\begin{itemize}
  \item When searching for related work, restrict your data sources to research papers. Do not use webpages or any other sources.
  \item All papers that you reference must be published on or before \texttt{cutoff\_date}. This is absolutely necessary. Double check every publication's date before referencing it. 
  \item The output format needs to be respected strictly. Do not change the format in any way.
  \item It is required to copy the in-text citations with their links in Markdown format \texttt{[(author, YYYY-MM)](link)} when referring to related work.   
\end{itemize}

\end{tcolorbox}

\begin{tcolorbox}[enhanced,breakable,title=Baselines Contribution Evaluation Prompt, label=box:baselines-contribution-1]

You are an expert research assistant. You are skilled at evaluating the novelty and impact of a research idea.

\medskip

First, your task is to identify a small number of high-level contribution dimensions. 
            
Contribution dimensions should represent general categories of scientific contribution that are meaningful and comparable across research ideas, regardless of the field.  
These might include:
\begin{itemize}
  \item \textbf{methodology} (e.g., proposing a new method, model, or procedure)
  \item \textbf{application} (e.g., applying existing methods to a new problem or domain)
  \item \textbf{theoretical contribution} (e.g., proving a new result, deriving a new model)
  \item \textbf{data} (e.g., constructing a new dataset, conducting original measurements or surveys)
  \item \textbf{evaluation} (e.g., designing an experimental protocol, benchmarking a technique)
  \item \textbf{tool or system design} (e.g., building software, devices, or infrastructure to support research)
  \item \textbf{conceptual framework} (e.g., introducing a new taxonomy or way of thinking about a problem)
\end{itemize}

Do not limit your output to the examples above but rather generate suitable dimensions for the research idea given to you. Only include dimensions that are actually reflected in the research idea — do not add generic or speculative categories. Please do not generate redundant dimensions.

\medskip

After having identified the dimensions, search through related literature for similar work and compare the novelty and impact of the research idea to these works along each dimension.  

Please summarize the results of your research into separate sections with biggest strengths, weaknesses, and suggestions for improvement according to the novelty and impact of the extracted dimensions compared to related work.

\medskip

It is required to put the in-text citations with their links in Markdown format \texttt{[(author, YYYY-MM)](link)} when referring to related work.

\medskip

Here is the research idea to evaluate:

[start research idea]  
\texttt{\{research\_idea\}}  
[end research idea]



\begin{lstlisting}[language=json]
<dimension_name_1>
"strengths": "direct comparison with literature where the 
research idea is MORE novel and/or impactful under this 
dimension",
"weaknesses": "direct comparison with literature where the 
research idea is LESS novel and/or impactful under this 
dimension",
"suggestions": "actionable and useful suggestions to improve 
the novelty and impact of the work, if needed, based on the 
evidence from the strengths and weaknesses sections"

<dimension_name_2> 
"strengths": "direct comparison with literature where the 
research idea is MORE novel and/or impactful under this 
dimension",
"weaknesses": "direct comparison with literature where the
research idea is LESS novel and/or impactful under this 
dimension",
"suggestions": "actionable and useful suggestions to improve 
the novelty and impact of the work, if needed, based on the 
evidence from the strengths and weaknesses sections"

<dimension_name_n> 
"strengths": "direct comparison with literature where the 
research idea is MORE novel and/or impactful under this 
dimension",
"weaknesses": "direct comparison with literature where the 
research idea is LESS novel and/or impactful under this 
dimension",
"suggestions": "actionable and useful suggestions to improve
the novelty and impact of the work, if needed, based on the 
evidence from the strengths and weaknesses sections"
\end{lstlisting}

\medskip

\textbf{Guidelines:}
\begin{itemize}
  \item When searching for related work, restrict your data sources to research papers. Do not use webpages or any other sources.
  \item All papers that you reference must be published on or before \texttt{\{cutoff\_date\}}. This is absolutely necessary. Double check every publication's date before referencing it. 
  \item The output format needs to be respected strictly. Do not change the format in any way.
  \item It is required to copy the in-text citations with their links in Markdown format \texttt{[(author, YYYY-MM)](link)} when referring to related work. Links need to be present.
\end{itemize}

\end{tcolorbox}
    
\subsection{\system}\label{app:instantiation}
We use the following parameter values to ensure a balance between the system performance, latency, and context window limit concerns:
\begin{itemize}
    \item Snippet search queries generated per extracted method: 1. 
    \item Snippets returned per query: up to 20. 
    \item Paper search queries generated per contribution statement: 3. 
    \item Papers returned per query: up to 20. 
    \item Papers returned by the recommendations for each seed paper: up to 8. 
    \item Papers returned via references augmentation for each seed paper: up to 10. 
    \item Papers sampled from the final list for pairwise comparison: up to 25. 
\end{itemize}
Additionally, we implement functionality that restricts the retrieved snippets and papers to those published prior to the cutoff date (i.e. the publication date of the paper that the research idea was extracted from). 
\subsection{Ablations}
In our ablation experiments, we use the same setup and parameters described in \autoref{app:instantiation}. 

To ablate the methods and results extraction (MRE), we remove the steps where we extract references from the snippets, download the papers, and extract the methods and results from each. Instead, the snippets are directly summarized and forwarded for final soundness review synthesis. 

The ablation of the paper augmentation (PA) is conducted by removing both the recommendation based augmentation and citation based augmentation. As such, only the seed list of papers identified after the initial paper search and relevance assessment are forwarded for pairwise comparison. 

Finally, to ablate the pairwise comparison (PC) step, we sample 25 papers from the final augmented paper list and forward them directly to the contribution review synthesis step. 
\section{Evaluation results}\label{app:results}
\subsection{Additional Results and Significance Analysis}
In this section, we include detailed evaluation results supplementing the results we presented in the main paper. Namely, Tables \ref{tab:pairwise_overall}, \ref{tab:pairwise_ai}, \ref{tab:pairwise_neuro}, \ref{tab:pairwise_biochem}, and \ref{tab:pairwise_ecology} show the pairwise significance analysis of the coverage results between \system variants and every baseline overall and for each discipline. Additionally, Table \ref{tab:type_axis_severity} shows the coverage results across all systems by type, axis, severity. Moreover, we conduct evaluation on \benchmark-AI using o4-mini (the underlying model of o4-mini-deep research) as the model backbone for \system, shown in Table \ref{tab:o4-mini-backbone}. These results showcase that using the same underlying model, \system still achieves higher coverage of the rubrics in \benchmark. We also conduct an evaluation of the related system that is closest in scope to the contribution module in \system, Idea Novelty Checker by \citep{shahid-etal-2025-literature}, which outputs novelty verdicts for ideas followed by short rationales (Table \ref{tab:idea-novelty-checker}). This evaluation contextualizes our system against other scholarly work on idea evaluation and confirms our choice to compare \system against strong baseline that output detailed feedback at the same scope as \system (i.e. strong LLMs and deep research systems). Finally, table \ref{tab:latency_cost} outlines the latency and cost of \system and baseline systems. 
\begin{table*}[h]
\small
\centering
\resizebox{\textwidth}{!}{
\begin{tabular}{llcccc}
\toprule
\textsc{ScholarEval} & Baseline & Mean$_{\text{SE}}$ (SD) & Mean$_{\text{BL}}$ (SD) & $p$ & Sig.? \\
\midrule
\system$_{\text{Llama}}$ & Llama-3.3-70B & 2.04 (1.16) & 1.83 (0.89) & $<$0.001 & Yes \\
 & GPT-4.1 & 2.04 (1.16) & 2.18 (1.08) & 0.998 & No \\
 & GPT-5.1 Instant & 2.04 (1.16) & 1.94 (0.98) & 0.015 & Yes \\
 & Claude-4-Sonnet & 2.04 (1.16) & 2.18 (1.04) & 0.998 & No \\
 & GPT-4o-search-preview & 2.04 (1.16) & 1.90 (0.98) & 0.001 & Yes \\
 & OpenAI Deep Research & 2.04 (1.16) & 2.28 (1.07) & 1.000 & No \\
 & DR Tulu & 2.04 (1.16) & 2.33 (1.14) & 1.000 & No \\
\midrule
\system$_{\text{GPT-4.1}}$ & Llama-3.3-70B & 2.72 (1.47) & 1.83 (0.89) & $<$0.001 & Yes \\
 & GPT-4.1 & 2.72 (1.47) & 2.18 (1.08) & $<$0.001 & Yes \\
 & GPT-5.1 Instant & 2.72 (1.47) & 1.94 (0.98) & $<$0.001 & Yes \\
 & Claude-4-Sonnet & 2.72 (1.47) & 2.18 (1.04) & $<$0.001 & Yes \\
 & GPT-4o-search-preview & 2.72 (1.47) & 1.90 (0.98) & $<$0.001 & Yes \\
 & OpenAI Deep Research & 2.72 (1.47) & 2.28 (1.07) & $<$0.001 & Yes \\
 & DR Tulu & 2.72 (1.47) & 2.33 (1.14) & $<$0.001 & Yes \\
\midrule
\system$_{\text{GPT-5.1}}$ & Llama-3.3-70B & 2.17 (1.27) & 1.83 (0.89) & $<$0.001 & Yes \\
 & GPT-4.1 & 2.17 (1.27) & 2.18 (1.08) & 0.578 & No \\
 & GPT-5.1 Instant & 2.17 (1.27) & 1.94 (0.98) & $<$0.001 & Yes \\
 & Claude-4-Sonnet & 2.17 (1.27) & 2.18 (1.04) & 0.579 & No \\
 & GPT-4o-search-preview & 2.17 (1.27) & 1.90 (0.98) & $<$0.001 & Yes \\
 & OpenAI Deep Research & 2.17 (1.27) & 2.28 (1.07) & 0.985 & No \\
 & DR Tulu & 2.17 (1.27) & 2.33 (1.14) & 0.999 & No \\
\midrule
\system$_{\text{Claude}}$ & Llama-3.3-70B & 2.77 (1.40) & 1.83 (0.89) & $<$0.001 & Yes \\
 & GPT-4.1 & 2.77 (1.40) & 2.18 (1.08) & $<$0.001 & Yes \\
 & GPT-5.1 Instant & 2.77 (1.40) & 1.94 (0.98) & $<$0.001 & Yes \\
 & Claude-4-Sonnet & 2.77 (1.40) & 2.18 (1.04) & $<$0.001 & Yes \\
 & GPT-4o-search-preview & 2.77 (1.40) & 1.90 (0.98) & $<$0.001 & Yes \\
 & OpenAI Deep Research & 2.77 (1.40) & 2.28 (1.07) & $<$0.001 & Yes \\
 & DR Tulu & 2.77 (1.40) & 2.33 (1.14) & $<$0.001 & Yes \\
\bottomrule
\end{tabular}
}
\caption{Pairwise comparisons for Overall coverage: Welch's t-tests comparing \system variants to baselines (n=1076 per model). Two-sided p-values shown; Sig.? indicates \system $>$ baseline with $p<0.05$.}
\label{tab:pairwise_overall}
\end{table*}

\begin{table*}[h]
\small
\centering
\resizebox{\textwidth}{!}{
\begin{tabular}{llcccc}
\toprule
\textsc{ScholarEval} & Baseline & Mean$_{\text{SE}}$ (SD) & Mean$_{\text{BL}}$ (SD) & $p$ & Sig.? \\
\midrule
\system$_{\text{Llama}}$ & Llama-3.3-70B & 2.06 (1.14) & 1.88 (0.92) & 0.006 & Yes \\
 & GPT-4.1 & 2.06 (1.14) & 2.21 (1.12) & 0.973 & No \\
 & GPT-5.1 Instant & 2.06 (1.14) & 2.09 (0.99) & 0.659 & No \\
 & Claude-4-Sonnet & 2.06 (1.14) & 2.28 (1.02) & 0.998 & No \\
 & GPT-4o-search-preview & 2.06 (1.14) & 1.95 (0.97) & 0.065 & No \\
 & OpenAI Deep Research & 2.06 (1.14) & 2.35 (1.07) & 1.000 & No \\
 & DR Tulu & 2.06 (1.14) & 2.35 (1.11) & 1.000 & No \\
\midrule
\system$_{\text{GPT-4.1}}$ & Llama-3.3-70B & 2.84 (1.39) & 1.88 (0.92) & $<$0.001 & Yes \\
 & GPT-4.1 & 2.84 (1.39) & 2.21 (1.12) & $<$0.001 & Yes \\
 & GPT-5.1 Instant & 2.84 (1.39) & 2.09 (0.99) & $<$0.001 & Yes \\
 & Claude-4-Sonnet & 2.84 (1.39) & 2.28 (1.02) & $<$0.001 & Yes \\
 & GPT-4o-search-preview & 2.84 (1.39) & 1.95 (0.97) & $<$0.001 & Yes \\
 & OpenAI Deep Research & 2.84 (1.39) & 2.35 (1.07) & $<$0.001 & Yes \\
 & DR Tulu & 2.84 (1.39) & 2.35 (1.11) & $<$0.001 & Yes \\
\midrule
\system$_{\text{GPT-5.1}}$ & Llama-3.3-70B & 2.31 (1.30) & 1.88 (0.92) & $<$0.001 & Yes \\
 & GPT-4.1 & 2.31 (1.30) & 2.21 (1.12) & 0.115 & No \\
 & GPT-5.1 Instant & 2.31 (1.30) & 2.09 (0.99) & 0.003 & Yes \\
 & Claude-4-Sonnet & 2.31 (1.30) & 2.28 (1.02) & 0.354 & No \\
 & GPT-4o-search-preview & 2.31 (1.30) & 1.95 (0.97) & $<$0.001 & Yes \\
 & OpenAI Deep Research & 2.31 (1.30) & 2.35 (1.07) & 0.688 & No \\
 & DR Tulu & 2.31 (1.30) & 2.35 (1.11) & 0.685 & No \\
\midrule
\system$_{\text{Claude}}$ & Llama-3.3-70B & 2.91 (1.34) & 1.88 (0.92) & $<$0.001 & Yes \\
 & GPT-4.1 & 2.91 (1.34) & 2.21 (1.12) & $<$0.001 & Yes \\
 & GPT-5.1 Instant & 2.91 (1.34) & 2.09 (0.99) & $<$0.001 & Yes \\
 & Claude-4-Sonnet & 2.91 (1.34) & 2.28 (1.02) & $<$0.001 & Yes \\
 & GPT-4o-search-preview & 2.91 (1.34) & 1.95 (0.97) & $<$0.001 & Yes \\
 & OpenAI Deep Research & 2.91 (1.34) & 2.35 (1.07) & $<$0.001 & Yes \\
 & DR Tulu & 2.91 (1.34) & 2.35 (1.11) & $<$0.001 & Yes \\
\bottomrule
\end{tabular}
}
\caption{Pairwise comparisons for AI coverage: Welch's t-tests comparing \system variants to baselines (n=425 per model). Two-sided p-values shown; Sig.? indicates \system $>$ baseline with $p<0.05$.}
\label{tab:pairwise_ai}
\end{table*}

\begin{table*}[h]
\small
\centering
\resizebox{\textwidth}{!}{
\begin{tabular}{llcccc}
\toprule
\textsc{ScholarEval} & Baseline & Mean$_{\text{SE}}$ (SD) & Mean$_{\text{BL}}$ (SD) & $p$ & Sig.? \\
\midrule
\system$_{\text{Llama}}$ & Llama-3.3-70B & 2.05 (1.18) & 1.82 (0.86) & 0.003 & Yes \\
 & GPT-4.1 & 2.05 (1.18) & 2.18 (1.10) & 0.923 & No \\
 & GPT-5.1 Instant & 2.05 (1.18) & 1.85 (0.99) & 0.011 & Yes \\
 & Claude-4-Sonnet & 2.05 (1.18) & 2.15 (1.08) & 0.866 & No \\
 & GPT-4o-search-preview & 2.05 (1.18) & 1.84 (0.96) & 0.007 & Yes \\
 & OpenAI Deep Research & 2.05 (1.18) & 2.25 (1.04) & 0.988 & No \\
 & DR Tulu & 2.05 (1.18) & 2.31 (1.17) & 0.997 & No \\
\midrule
\system$_{\text{GPT-4.1}}$ & Llama-3.3-70B & 2.74 (1.55) & 1.82 (0.86) & $<$0.001 & Yes \\
 & GPT-4.1 & 2.74 (1.55) & 2.18 (1.10) & $<$0.001 & Yes \\
 & GPT-5.1 Instant & 2.74 (1.55) & 1.85 (0.99) & $<$0.001 & Yes \\
 & Claude-4-Sonnet & 2.74 (1.55) & 2.15 (1.08) & $<$0.001 & Yes \\
 & GPT-4o-search-preview & 2.74 (1.55) & 1.84 (0.96) & $<$0.001 & Yes \\
 & OpenAI Deep Research & 2.74 (1.55) & 2.25 (1.04) & $<$0.001 & Yes \\
 & DR Tulu & 2.74 (1.55) & 2.31 (1.17) & $<$0.001 & Yes \\
\midrule
\system$_{\text{GPT-5.1}}$ & Llama-3.3-70B & 2.12 (1.24) & 1.82 (0.86) & $<$0.001 & Yes \\
 & GPT-4.1 & 2.12 (1.24) & 2.18 (1.10) & 0.739 & No \\
 & GPT-5.1 Instant & 2.12 (1.24) & 1.85 (0.99) & 0.001 & Yes \\
 & Claude-4-Sonnet & 2.12 (1.24) & 2.15 (1.08) & 0.627 & No \\
 & GPT-4o-search-preview & 2.12 (1.24) & 1.84 (0.96) & $<$0.001 & Yes \\
 & OpenAI Deep Research & 2.12 (1.24) & 2.25 (1.04) & 0.922 & No \\
 & DR Tulu & 2.12 (1.24) & 2.31 (1.17) & 0.976 & No \\
\midrule
\system$_{\text{Claude}}$ & Llama-3.3-70B & 2.55 (1.45) & 1.82 (0.86) & $<$0.001 & Yes \\
 & GPT-4.1 & 2.55 (1.45) & 2.18 (1.10) & $<$0.001 & Yes \\
 & GPT-5.1 Instant & 2.55 (1.45) & 1.85 (0.99) & $<$0.001 & Yes \\
 & Claude-4-Sonnet & 2.55 (1.45) & 2.15 (1.08) & $<$0.001 & Yes \\
 & GPT-4o-search-preview & 2.55 (1.45) & 1.84 (0.96) & $<$0.001 & Yes \\
 & OpenAI Deep Research & 2.55 (1.45) & 2.25 (1.04) & 0.002 & Yes \\
 & DR Tulu & 2.55 (1.45) & 2.31 (1.17) & 0.011 & Yes \\
\bottomrule
\end{tabular}
}
\caption{Pairwise comparisons for Neuroscience coverage: Welch's t-tests comparing \system variants to baselines (n=314 per model). Two-sided p-values shown; Sig.? indicates \system $>$ baseline with $p<0.05$.}
\label{tab:pairwise_neuro}
\end{table*}

\begin{table*}[h]
\small
\centering
\resizebox{\textwidth}{!}{
\begin{tabular}{llcccc}
\toprule
\textsc{ScholarEval} & Baseline & Mean$_{\text{SE}}$ (SD) & Mean$_{\text{BL}}$ (SD) & $p$ & Sig.? \\
\midrule
\system$_{\text{Llama}}$ & Llama-3.3-70B & 2.04 (1.19) & 1.81 (0.90) & 0.031 & Yes \\
 & GPT-4.1 & 2.04 (1.19) & 2.14 (1.04) & 0.778 & No \\
 & GPT-5.1 Instant & 2.04 (1.19) & 1.68 (0.91) & 0.002 & Yes \\
 & Claude-4-Sonnet & 2.04 (1.19) & 2.01 (1.01) & 0.408 & No \\
 & GPT-4o-search-preview & 2.04 (1.19) & 1.86 (1.00) & 0.081 & No \\
 & OpenAI Deep Research & 2.04 (1.19) & 2.08 (1.06) & 0.619 & No \\
 & DR Tulu & 2.04 (1.19) & 2.24 (1.04) & 0.937 & No \\
\midrule
\system$_{\text{GPT-4.1}}$ & Llama-3.3-70B & 2.61 (1.42) & 1.81 (0.90) & $<$0.001 & Yes \\
 & GPT-4.1 & 2.61 (1.42) & 2.14 (1.04) & $<$0.001 & Yes \\
 & GPT-5.1 Instant & 2.61 (1.42) & 1.68 (0.91) & $<$0.001 & Yes \\
 & Claude-4-Sonnet & 2.61 (1.42) & 2.01 (1.01) & $<$0.001 & Yes \\
 & GPT-4o-search-preview & 2.61 (1.42) & 1.86 (1.00) & $<$0.001 & Yes \\
 & OpenAI Deep Research & 2.61 (1.42) & 2.08 (1.06) & $<$0.001 & Yes \\
 & DR Tulu & 2.61 (1.42) & 2.24 (1.04) & 0.006 & Yes \\
\midrule
\system$_{\text{GPT-5.1}}$ & Llama-3.3-70B & 1.96 (1.19) & 1.81 (0.90) & 0.112 & No \\
 & GPT-4.1 & 1.96 (1.19) & 2.14 (1.04) & 0.916 & No \\
 & GPT-5.1 Instant & 1.96 (1.19) & 1.68 (0.91) & 0.012 & Yes \\
 & Claude-4-Sonnet & 1.96 (1.19) & 2.01 (1.01) & 0.651 & No \\
 & GPT-4o-search-preview & 1.96 (1.19) & 1.86 (1.00) & 0.218 & No \\
 & OpenAI Deep Research & 1.96 (1.19) & 2.08 (1.06) & 0.819 & No \\
 & DR Tulu & 1.96 (1.19) & 2.24 (1.04) & 0.984 & No \\
\midrule
\system$_{\text{Claude}}$ & Llama-3.3-70B & 2.64 (1.40) & 1.81 (0.90) & $<$0.001 & Yes \\
 & GPT-4.1 & 2.64 (1.40) & 2.14 (1.04) & $<$0.001 & Yes \\
 & GPT-5.1 Instant & 2.64 (1.40) & 1.68 (0.91) & $<$0.001 & Yes \\
 & Claude-4-Sonnet & 2.64 (1.40) & 2.01 (1.01) & $<$0.001 & Yes \\
 & GPT-4o-search-preview & 2.64 (1.40) & 1.86 (1.00) & $<$0.001 & Yes \\
 & OpenAI Deep Research & 2.64 (1.40) & 2.08 (1.06) & $<$0.001 & Yes \\
 & DR Tulu & 2.64 (1.40) & 2.24 (1.04) & 0.003 & Yes \\
\bottomrule
\end{tabular}
}
\caption{Pairwise comparisons for Biochemistry coverage: Welch's t-tests comparing \system variants to baselines (n=147 per model). Two-sided p-values shown; Sig.? indicates \system $>$ baseline with $p<0.05$.}
\label{tab:pairwise_biochem}
\end{table*}

\begin{table*}[h]
\small
\centering
\resizebox{\textwidth}{!}{
\begin{tabular}{llcccc}
\toprule
\textsc{ScholarEval} & Baseline & Mean$_{\text{SE}}$ (SD) & Mean$_{\text{BL}}$ (SD) & $p$ & Sig.? \\
\midrule
\system$_{\text{Llama}}$ & Llama-3.3-70B & 1.94 (1.13) & 1.74 (0.87) & 0.027 & Yes \\
 & GPT-4.1 & 1.94 (1.13) & 2.13 (1.00) & 0.958 & No \\
 & GPT-5.1 Instant & 1.94 (1.13) & 1.95 (0.97) & 0.537 & No \\
 & Claude-4-Sonnet & 1.94 (1.13) & 2.11 (1.03) & 0.937 & No \\
 & GPT-4o-search-preview & 1.94 (1.13) & 1.95 (1.02) & 0.536 & No \\
 & OpenAI Deep Research & 1.94 (1.13) & 2.35 (1.11) & 1.000 & No \\
 & DR Tulu & 1.94 (1.13) & 2.39 (1.20) & 1.000 & No \\
\midrule
\system$_{\text{GPT-4.1}}$ & Llama-3.3-70B & 2.52 (1.51) & 1.74 (0.87) & $<$0.001 & Yes \\
 & GPT-4.1 & 2.52 (1.51) & 2.13 (1.00) & 0.002 & Yes \\
 & GPT-5.1 Instant & 2.52 (1.51) & 1.95 (0.97) & $<$0.001 & Yes \\
 & Claude-4-Sonnet & 2.52 (1.51) & 2.11 (1.03) & 0.001 & Yes \\
 & GPT-4o-search-preview & 2.52 (1.51) & 1.95 (1.02) & $<$0.001 & Yes \\
 & OpenAI Deep Research & 2.52 (1.51) & 2.35 (1.11) & 0.106 & No \\
 & DR Tulu & 2.52 (1.51) & 2.39 (1.20) & 0.177 & No \\
\midrule
\system$_{\text{GPT-5.1}}$ & Llama-3.3-70B & 2.09 (1.28) & 1.74 (0.87) & $<$0.001 & Yes \\
 & GPT-4.1 & 2.09 (1.28) & 2.13 (1.00) & 0.633 & No \\
 & GPT-5.1 Instant & 2.09 (1.28) & 1.95 (0.97) & 0.115 & No \\
 & Claude-4-Sonnet & 2.09 (1.28) & 2.11 (1.03) & 0.567 & No \\
 & GPT-4o-search-preview & 2.09 (1.28) & 1.95 (1.02) & 0.120 & No \\
 & OpenAI Deep Research & 2.09 (1.28) & 2.35 (1.11) & 0.982 & No \\
 & DR Tulu & 2.09 (1.28) & 2.39 (1.20) & 0.991 & No \\
\midrule
\system$_{\text{Claude}}$ & Llama-3.3-70B & 2.90 (1.39) & 1.74 (0.87) & $<$0.001 & Yes \\
 & GPT-4.1 & 2.90 (1.39) & 2.13 (1.00) & $<$0.001 & Yes \\
 & GPT-5.1 Instant & 2.90 (1.39) & 1.95 (0.97) & $<$0.001 & Yes \\
 & Claude-4-Sonnet & 2.90 (1.39) & 2.11 (1.03) & $<$0.001 & Yes \\
 & GPT-4o-search-preview & 2.90 (1.39) & 1.95 (1.02) & $<$0.001 & Yes \\
 & OpenAI Deep Research & 2.90 (1.39) & 2.35 (1.11) & $<$0.001 & Yes \\
 & DR Tulu & 2.90 (1.39) & 2.39 (1.20) & $<$0.001 & Yes \\
\bottomrule
\end{tabular}
}
\caption{Pairwise comparisons for Ecology coverage: Welch's t-tests comparing \system variants to baselines (n=190 per model). Two-sided p-values shown; Sig.? indicates \system $>$ baseline with $p<0.05$.}
\label{tab:pairwise_ecology}
\end{table*}

\begin{table*}[h]
\centering
\resizebox{\textwidth}{!}{
\begin{tabular}{l cc cc cc}
\toprule
\textbf{System} & \multicolumn{2}{c}{\textbf{Type}} & \multicolumn{2}{c}{\textbf{Axis}} & \multicolumn{2}{c}{\textbf{Severity}} \\
\cmidrule(lr){2-3} \cmidrule(lr){4-5} \cmidrule(lr){6-7}
 & Strength & Weakness & Soundness & Contribution & Major & Minor \\
\midrule
Llama-3.3-70B          & 2.63 (0.82) & 1.64 (0.80) & 1.72 (0.84) & 2.14 (0.95) & 1.90 (0.88) & 1.72 (0.90) \\
GPT-4.1                & 3.21 (1.05) & 1.94 (0.95) & 2.00 (0.99) & 2.71 (1.16) & 2.28 (1.05) & 2.01 (1.12) \\
Claude-4-Sonnet        & 3.04 (0.97) & 1.98 (0.95) & 2.06 (1.01) & 2.51 (1.08) & 2.28 (1.03) & 2.01 (1.04) \\
GPT-4o-search-preview  & 2.83 (0.98) & 1.69 (0.85) & 1.76 (0.91) & 2.32 (1.05) & 1.99 (0.98) & 1.77 (0.98) \\
OpenAI Deep Research  & 2.98 (0.90) & 2.12 (1.04) & 2.17 (1.05) & 2.60 (1.05) & 2.41 (1.05) & 2.08 (1.06) \\
\midrule
\system$_{\text{Llama}}$      & 2.93 (1.19) & 1.83 (1.05) & 1.89 (1.10) & 2.46 (1.21) & 2.16 (1.14) & 1.83 (1.15) \\
\system$_{\text{GPT}}$        & 4.16 (1.07) & 2.40 (1.34) & 2.51 (1.43) & 3.36 (1.40) & 2.92 (1.44) & 2.41 (1.45) \\
\system$_{\text{Claude}}$     & 3.65 (1.21) & 2.56 (1.36) & 2.61 (1.37) & 3.24 (1.37) & 2.93 (1.38) & 2.51 (1.39) \\
\bottomrule
\end{tabular}
}
\caption{Coverage results of different variants of \system and baselines across type, axis, and severity dimensions (mean scores with standard deviations).}
\label{tab:type_axis_severity}
\end{table*}

\begin{table*}[h]
\small
\centering
\begin{tabular}{l c}
\hline
\textbf{System} & \textbf{Coverage (mean $\pm$ std)} \\
\hline
OpenAI Deep Research & $2.35 \pm 1.07$ \\
\system$_{\text{GPT}}$ & $2.84 \pm 1.39$ \\
\system$_{\text{Claude}}$ & $2.91 \pm 1.34$ \\
\system$_{\text{o4-mini}}$ & $2.78 \pm 1.35$ \\
\hline
\end{tabular}
\caption{\system coverage on \benchmark-AI using o4-mini as backbone}
\label{tab:o4-mini-backbone}
\end{table*}

\begin{table*}[h]
\small
\centering
\begin{tabular}{l c}
\hline
\textbf{System} & \textbf{Coverage (mean (std))} \\
\hline
Llama-3.3-70B & 2.22 (1.04) \\
GPT-4.1 & 2.78 (1.15) \\
Claude-4-Sonnet & 2.57 (1.05) \\
GPT-4o-search-preview & 2.36 (1.05) \\
OpenAI Deep Research & 2.66 (1.06) \\
ScholarEval-Llama-3.3 & 2.52 (1.17) \\
ScholarEval-GPT-4.1 & 3.45 (1.34) \\
ScholarEval-Claude-4-Sonnet & 3.34 (1.34) \\
Idea Novelty Checker \citep{shahid-etal-2025-literature} & 2.27 (1.15) \\
\hline
\end{tabular}
\caption{Coverage of contribution rubrics in \benchmark-AI of Idea Novelty Checker \citep{shahid-etal-2025-literature}}
\label{tab:idea-novelty-checker}
\end{table*}
\begin{table*}[h]
\centering
\small
\begin{tabular}{lcc}
\toprule
\textbf{System} & \textbf{Latency (mins)} & \textbf{Cost (USD)} \\
\midrule
Llama-3.3-70B         & 1.23 & 0.00 \\
GPT-4.1               & 2.90 & 0.03 \\
Claude-4-Sonnet       & 7.06 & 0.08 \\
GPT-4o-search-preview & 1.94 & 0.15 \\
OpenAI Deep Research & 3.33 & 0.49 \\
ScholarEval-Llama     & 6.63 & 0.00 \\
ScholarEval-GPT       & 10.23 & 2.54 \\
ScholarEval-Claude    & 12.10 & 3.38 \\
\bottomrule
\end{tabular}
\caption{Latency and cost per run (soundness and contribution) for \system and baselines.}
\label{tab:latency_cost}
\end{table*}

\subsection{Manual Audit}\label{app:manual-audit}
Our manual audit of papers referenced in the evaluation reports generated by baseline systems underscores various failure modes that undermine the reliability of these systems for literature-grounded research idea evaluation. 

First, our inspection of the links indicates a much high reference invalidity rate than the conservative lowerbound reported in \Sref{sec:results}. For Llama-3.3-70B, some our inspected reports had up to 90\% invalid references, and stronger LLMs such as Claude 4 Sonnet had up to 50\% reference invalidity. We also observed that this issue is not eliminated by retrieval, as 22\% of the links included in an inspected report generated by GPT-4o-search-preview were invalid. 

We have also noticed subtler failure modes. In the example below, Claude 4 Sonnet references a paper by Gong et al. However, upon inspection, we notice that the linked paper is in fact authored by \citet{Bakalarski2016ISDetect}.
\begin{quote}
While [(Gong et al., 2016-08)](\url{https://www.nature.com/articles/nbt.3621}) demonstrated DNA-barcoded antibodies for multiplexed imaging, their approach relied on conventional chemical conjugation methods that can compromise antibody function.
\end{quote}
These errors in attribution are also made by GPT-4o-search-preview. In the example below, it wrongly attributes a publication by \citet{Brotherton2013ColibactinProdrug} to Kazane et al.  
\begin{quote}
These techniques have been widely used to evaluate the impact of conjugation on antibody performance, ensuring that the modifications do not adversely affect antigen-binding capabilities [(Kazane et al., 2013)](\url{https://pubs.acs.org/doi/10.1021/ja312154m}).
\end{quote}
We also observed that o4-mini-deep-research commits similar attribution errors. In the example below, it attributes CyCIF \citep{Lin2016CycIF} and CODEX \citep{Kuswanto2023CODEX} to the wrong authors. 
\begin{quote}
The proposed system could show improved signal (via HCR) and straightforward reagent generation (via MaMBA). The plan’s demonstration of 12-plex imaging is on par with existing methods like CyCIF (Gerdes et al. 2013) or CODEX (Goltsev et al. 2018), indicating strong competitive impact.
\end{quote}
These observations showcase the limitations of even strong baselines (web-connected LMs and deep research systems) in generating reliable literature-backed research idea evaluations. 
\section{Additional Details from the Expert User Study}
\label{app:userstudy}
\subsection{Setup and Materials}
\textbf{Rubric}
We include our full rubric for the user study in \autoref{tab:user_study_questions}. Answers were collected via Google Forms.
\begin{table*}[ht]
\centering
\resizebox{\textwidth}{!}{
\begin{tabular}{lcp{3cm}p{8cm}}
\toprule
Dimension & Module & Scale & Question \\
\midrule
\metricpillorange{\small IdeaFaithful} & Soundness & 1-10 & The number of extracted methods/experiments that are faithful to my original intention \\
\metricpillorange{\small IdeaFaithful} & Contribution & 1-10 & The number of dimensions that are faithful to my original intention \\

\metricpillorange{\small Focus} & Soundness/Contribution & 1-10 & The strengths reinforce the most valuable aspects of the proposal \\
\metricpillorange{\small Focus} & Soundness/Contribution & 1-10 & The weaknesses highlight the most deficient aspects of the proposal \\
\metricpillorange{\small Focus} & Soundness/Contribution & 1-10 & The suggestions for improvement are the most important aspects to address \\
\metricpillorange{\small Focus} & Contribution & 1-10 & The criteria used to compare your proposal to related work are based on relevant metrics and/or frameworks and are NOT arbitrary \\

\metricpillorange{\small Refine} & Soundness/Contribution & 1-10 & The top suggestions for improvement offer valuable, targeted, feasible modifications to your experiment \\

\metricpillorange{\small LitEngage} & Soundness/Contribution & 1-10 & The evaluation uses detailed comparisons with specific components of relevant literature, rather than providing superficial citations \\

\metricpillorange{\small Citations} & Soundness/Contribution & YYYY & Please estimate how many citations from your MAIN FIELD of study were new to you and you would use in your work \\
\metricpillorange{\small Citations} & Soundness/Contribution & YYYY & Please estimate how many citations from OUTSIDE  your main field of study were new to you and you would use in your work\\

\metricpillorange{\small Useful} & Soundness/Contribution & 1-10 & I found the [Module] evaluation useful \\
\metricpillorange{\small Useful} & Soundness/Contribution & 1-10 & I would use the [Module] evaluation again in the future \\
\bottomrule
\end{tabular}
}
\caption{Questions asked during the expert user study, and their respective dimension, module, and scale.}
\label{tab:user_study_questions}
\end{table*}

\textbf{Recruitment, Demographics, and Compensation.} Participants were recruited via X (Twitter) and graduate department emailing lists internationally. \autoref{tab:user_study_demographics} shows the number of evaluations completed per discipline, as well as years of experience. Experts were paid $\$25$ per research idea evaluated, with a bonus $\$10$ awarded if they completed written feedback.
\begin{table*}[ht]
\centering
\begin{tabular}{lcccc}
\toprule
Domain & Experts & Evaluations & 2-4 YoE & 4+ YoE \\
\midrule
Computer Science & 8 & 21 & 5 & 16 \\
Neuroscience & 4 & 12 & 6 & 6 \\
Chemistry & 4 & 8 & 1 & 7 \\
Ecology & 2 & 5 & 4 & 1 \\
Total & \textbf{18} & \textbf{46} & 16 & 30 \\
\bottomrule
\end{tabular}
\caption{User study demographics grouped by domain. This includes number of experts, evaluations, and years of experience per discipline. }
\label{tab:user_study_demographics}
\end{table*}

\begin{figure*}[h]  
    \centering
    \includegraphics[width=1\linewidth]{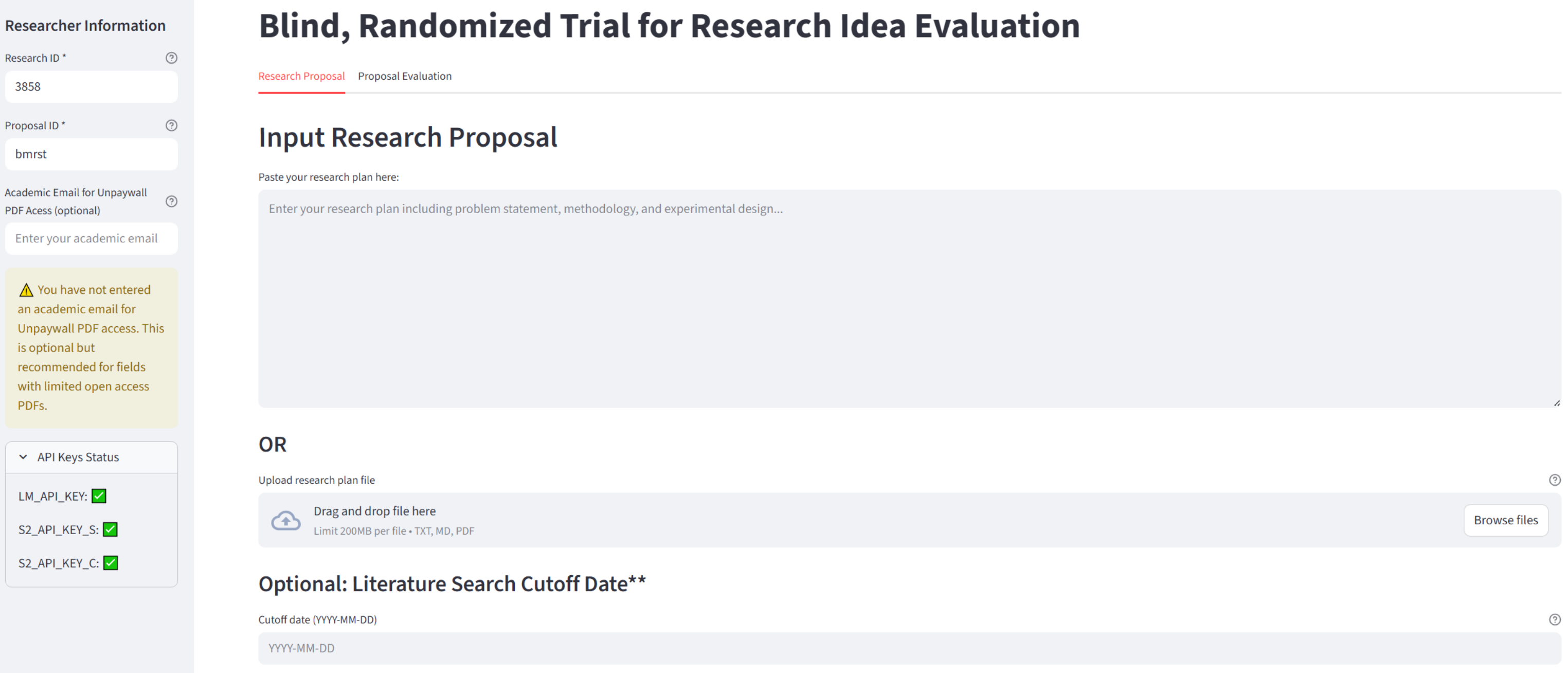}
    \caption{User interface for the Expert User Study. The landing page includes inputs for a research id, proposal id, email for unpaywall, research proposal text box or file upload, and a literature cutoff date.}
    \label{fig:us_ui}
\end{figure*}

\begin{figure*}[h]  
    \centering
    \includegraphics[width=1\linewidth]{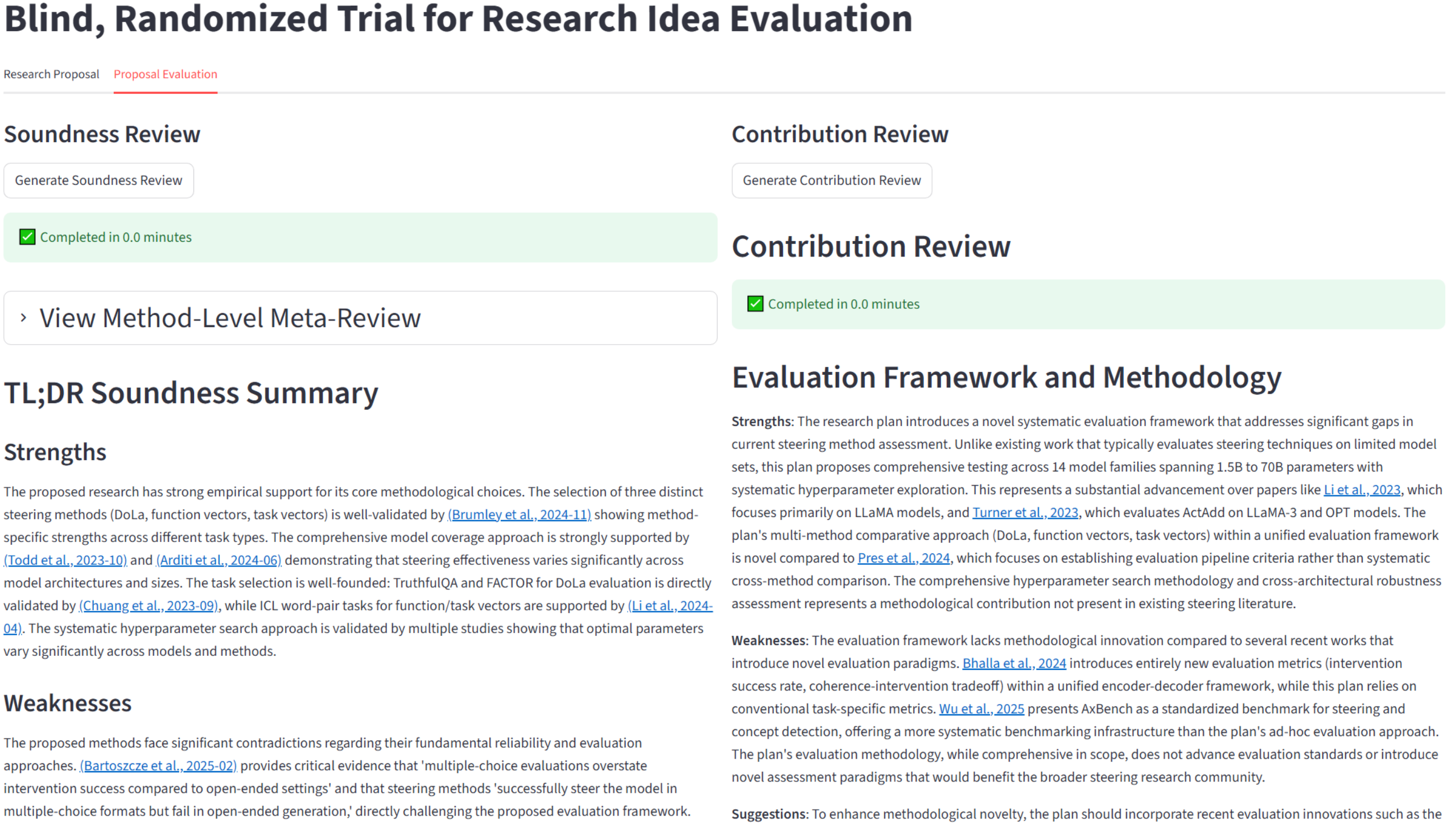}
    \caption{User interface for the Expert User Study, with an example from a user who gave permission to share their output. Reviews for both soundness and contribution are displayed in markdown format. There is a collapsible text box to show the method-level evaluations.}
    \label{fig:us_ui2}
\end{figure*}

\textbf{User Interface.} Experts were shown the interface shown in \autoref{fig:us_ui} and \autoref{fig:us_ui2}. We give experts a unique research id along with 4 unique idea keys, one for each research idea. Each is linked to a replicable, semi-random, and pre-calculated order of systems. The semi-random component comes from the fact that we forced the random order to include at least 2 of each system to ensure each person had the opportunity to contribute evaluations on either system. The script to regenerate these keys will be released with our code. We ensure validity of system blindness by asking users to guess which system they are evaluating, which has a $-0.12$ pearson correlation ($p=0.44$) with the actual assigned system.
\subsection{Statistical Methods}
\label{app:user_study_stats}
Because we collect multiple research ideas per person, our data is no longer independent, and we cannot use a mean-differences t-test. Instead, we use a Linear Mixed-Effects Model \citep{RaudenbushBryk2002}, which models both fixed effects (\system vs deep research) and random effects (ideas within and across participants). This helps to stabilize the ratings across our experts, as some may be consistently higher or lower raters. This additionally helps to account for some variability from resarch idea quality. The Linear Mixed-Effects Model is defined as follows:
\begin{equation}
\label{eq:mixed_effects}
\mathbf{y} = \mathbf{X}\boldsymbol{\beta} + \mathbf{Z}\mathbf{b} + \boldsymbol{\varepsilon}
\end{equation}

This adds an additional term, $\mathbf{Z}$, on top of the standard Linear Model to capture the variance of random effects, where $\mathbf{Z}$ is a matrix of shape $n{\times}m$, where $n$ is the total observations and $m$ is the number of unique participants. Each row is one-hot encoded for the participant who contributed the data point.
 

\section{\system output examples}\label{app:examples}
In this section we provide an example of a neuroscience research idea from \benchmark along with an excerpt of a method soundness review and contribution dimension generated by \system. 
\begin{tcolorbox}[enhanced,breakable,title=Research Idea: Neuroscience Research Idea, label=box:neuro-idea]

\textbf{Problem}  

We aim to investigate cortical plasticity at its boundaries by studying information processing in individuals who lost vision at birth or early in life. Total loss of a sensory modality provides a unique opportunity to characterize plasticity at its limits.  

Prior work shows that cortical structures activated by vision in sighted brains are recruited for cognitive functions, including braille reading, in visually deprived brains. However, overlap of functional responses alone does not reveal what information these activations represent or their cognitive role.  

We will study tactile braille reading in blind participants to clarify the transformation of sensory (hand-dependent) to perceptual (hand-independent) braille letter representations.

\medskip
\textbf{Method}  

We will combine fMRI and EEG in a multivariate framework to determine the cortical location and temporal emergence of sensory and perceptual representations, relating them to behavioral similarity ratings.  

Participants will read braille letters with either hand via piezo-electric refreshable cells, preventing finger movement artifacts. The stimulus set will include ten braille letters (eight for analysis, two as vigilance targets).  

\textit{Multivariate Classification:}  
\begin{itemize}
  \item fMRI voxel patterns $\rightarrow$ spatial location of sensory representations  
  \item EEG electrode patterns $\rightarrow$ temporal dynamics of representations  
  \item Perceptual representations: across-hand classification (train on one hand, test on other)  
  \item Sensory representations: within-hand classification minus across-hand classification  
\end{itemize}

\medskip
\textbf{Experiment Design}  

\textit{Participants and Stimuli:}  
Blind individuals (vision loss $\leq$3 years) will take part in:  
\begin{itemize}
  \item fMRI (N=15)  
  \item EEG (N=11)  
  \item Behavioral similarity ratings (N=19)  
\end{itemize}
Letters (B, C, D, L, M, N, V, Z) presented to left/right index fingers; E, O serve as catch stimuli.

\textit{fMRI:}  
500ms letter presentations, 2500ms ISI, 16 conditions (8 letters × 2 hands), repeated per run with catch and null trials. A localizer (letters vs. fake letters, both hands) will define ROIs along tactile and sighted reading pathways.

\textit{EEG:}  
Similar design with 500ms presentations, 500ms ISI, longer catch-trial intervals. 64 channels, standard 10-10 placement.

\textit{Behavioral:}  
Participants will rate perceived similarity of braille letter pairs (1=similar, 7=different) using adjacent braille cells.

\medskip
\textbf{Analysis Plan}  

We will focus on tactile areas (S1, S2, intra-parietal cortex, insula) and sighted reading areas (early visual cortex, V4, lateral occipital complex, letter form area, VWFA).  

\textit{Hypotheses:}  
\begin{enumerate}
  \item Sensory representations in tactile processing areas; perceptual representations in sighted reading areas.  
  \item Sensory representations emerge earlier in time than perceptual representations.  
  \item Representations correlate with behavioral similarity ratings, showing behavioral relevance.  
\end{enumerate}

\textit{Analysis Methods:}  
\begin{itemize}
  \item Region-of-interest and searchlight analyses for fMRI  
  \item Time-resolved classification for EEG  
  \item Non-parametric statistical tests with multiple comparison correction  
\end{itemize}

\end{tcolorbox}

\newpage
\begin{tcolorbox}[enhanced,breakable,title = \system Example of a Method Soundness Evaluation, label = box:neuro-soundness-example]

[...More method-level reviews...]

\textbf{Method: To isolate sensory representations from perceptual ones, we will use a two-step procedure for sensory representations: first conducting within-hand classification (training and testing on the same hand), then subtracting across-hand from within-hand classification results to isolate sensory from perceptual representations.}

\textit{\textbf{Support}}

The proposed two-step subtraction method (within-hand minus across-hand classification) receives strong empirical validation from multiple studies that successfully implemented nearly identical approaches. \href{https://www.semanticscholar.org/paper/05bd005b171e9196dc63e4ca177e05261b9eceb8}{(Pietrini et al., 2004-04)} provides the most direct precedent, employing the exact same methodology with multivariate pattern analysis using support vector machine classifiers to distinguish sensory from perceptual tactile representations in blind participants using refreshable braille cells. Their results demonstrated that within-hand classification captured both sensory and perceptual information while across-hand classification isolated perceptual representations, with the subtraction revealing hand-specific sensory information primarily in somatosensory cortices. \href{https://www.semanticscholar.org/paper/016b2410f1bc4753ba3339aa5caf3ffff5123dd7}{(Merabet et al., 2008-08)} corroborates this approach with quantitative validation, showing within-hand classification accuracies of 65-72\% versus across-hand accuracies of 55-62\% in somatosensory regions, yielding meaningful difference scores of 8-15\% that were functionally validated through TMS disruption experiments causing 12-18\% decreases in tactile discrimination performance. \href{https://www.semanticscholar.org/paper/6461bbc4235d83402303d14ba112ce0e1d24330e}{(Grigoryan et al., 2020-06)} extends this validation to the temporal domain using EEG, demonstrating that the within-hand versus across-hand distinction emerges early (100-200ms) for sensory representations and later for perceptual ones, directly supporting the proposed method's theoretical foundation. The methodological framework receives additional support from \href{https://www.semanticscholar.org/paper/43b3e0e6f30cbec4a6f24b1a89f4396425c0b054}{(Amedi et al., 2002-11)}, which successfully used cross-modal classification to isolate shared representational content, achieving significant above-chance performance in lateral occipital regions and demonstrating the feasibility of using classification approaches to dissociate different types of neural representations. The technical implementation is well-supported by multiple studies using refreshable braille cells with fMRI, including \href{https://www.semanticscholar.org/paper/43fdd133260f88d3c8b1d200b6c53ef49bb3a241}{(Burton et al., 2002-12)} and \href{https://www.semanticscholar.org/paper/78576ee83c0e47036b966c284f5b86469f9cb520}{(Kim et al., 2017-11)}, confirming the viability of piezo-electric braille stimulation during neuroimaging. The theoretical rationale is strengthened by extensive evidence for cross-modal plasticity in blind individuals, with \href{https://www.semanticscholar.org/paper/4999e7c7159b09e45dc37d4750a8d4dc94143113}{(Büchel et al., 1998-03)} and \href{https://www.semanticscholar.org/paper/b5456faa31cd76d7d0216173084d14daf38975ec}{(Sadato et al., 1998-07)} demonstrating that visual cortical areas are recruited for tactile processing in blind individuals, providing the neurobiological foundation for expecting both sensory and perceptual representations in repurposed visual regions.


\textbf{\textit{Contradictions}}

The proposed subtraction method faces several significant methodological and theoretical challenges that could undermine its validity. Most critically, the fundamental assumption that within-hand classification captures both sensory and perceptual information while across-hand classification captures only perceptual information may be oversimplified and potentially incorrect. The subtraction logic assumes these components are linearly additive and independent, but neural representations are likely to interact nonlinearly, making simple subtraction mathematically invalid. \href{https://www.semanticscholar.org/paper/45ae04ee78db50ba1a2454915b11090186a7d70d}{(Rothlein, and Rapp, 2014-04)}demonstrates more sophisticated approaches to isolating specific representational features using selectivity analyses, ANOVA comparisons, and regression approaches that account for multiple interacting factors simultaneously, suggesting the proposed binary subtraction approach may be too crude. The method assumes that sensory representations are purely hand-specific and perceptual representations are purely hand-independent, but this dichotomy may not reflect the true complexity of neural coding. Intermediate representations could exist that are partially hand-dependent due to factors like hand-specific motor learning, tactile sensitivity differences, or hemispheric processing asymmetries that are not purely sensory. \href{https://www.semanticscholar.org/paper/863c4090e64c43299bb825b3897c9edcf5dfee62}{(Grant et al.)} illustrates how discrimination thresholds vary systematically across conditions, suggesting that what appears to be 'sensory' versus 'perceptual' may actually reflect a continuum of processing stages rather than discrete categories. The statistical validity is questionable because the subtraction of two noisy classification accuracies will amplify noise and reduce statistical power, potentially leading to unreliable difference scores. \href{https://www.semanticscholar.org/paper/947258aa49e5079accd74e7530e7a9c0fd52358a}{(Siuda-Krzywicka et al., 2016-03)} used correlation-based representational similarity analysis with proper statistical controls (F(2,56)=14.53, p<0.001), demonstrating more rigorous approaches to comparing representational similarities that account for statistical dependencies. The method lacks validation against ground truth - there is no independent way to verify that the residual from subtraction actually represents 'pure' sensory information rather than artifacts, noise, or other confounding factors. The approach also fails to account for potential differences in classification difficulty between within-hand and across-hand conditions that could arise from factors unrelated to sensory versus perceptual distinctions, such as differences in signal-to-noise ratios, statistical power, or systematic biases in the classification algorithms.

\textbf{\textit{Suggested Action}}

The proposed method requires substantial methodological improvements to address its fundamental limitations. First, implement multiple complementary analysis approaches rather than relying solely on subtraction, including the selectivity analyses demonstrated by \href{https://www.semanticscholar.org/paper/45ae04ee78db50ba1a2454915b11090186a7d70d}{(Rothlein, and Rapp, 2014-04)} such as ANOVA selectivity comparisons and regression analyses that can isolate specific representational features while controlling for confounding factors. Second, establish proper statistical validation by implementing permutation testing with scrambled labels (as in the 1000-iteration approach used by \href{https://www.semanticscholar.org/paper/016b2410f1bc4753ba3339aa5caf3ffff5123dd7}{(Merabet et al., 2008-08)}) to determine chance-level performance for the subtraction scores, and use cluster-based correction for multiple comparisons as demonstrated by \href{https://www.semanticscholar.org/paper/6461bbc4235d83402303d14ba112ce0e1d24330e}{(Grigoryan et al., 2020-06)}. Third, incorporate representational similarity analysis following \href{https://www.semanticscholar.org/paper/947258aa49e5079accd74e7530e7a9c0fd52358a}{(Siuda-Krzywicka et al., 2016-03)} to examine correlations between neural patterns and behavioral similarity ratings, providing independent validation of the proposed sensory versus perceptual distinction. Fourth, add control analyses to test the method's assumptions, including examining whether the subtraction approach yields consistent results when applied to different classification algorithms, testing whether intermediate levels of hand-dependence exist by examining classification performance across different finger positions or stimulation intensities, and validating that observed differences are not due to systematic biases in classification difficulty. Fifth, implement functional validation similar to \href{https://www.semanticscholar.org/paper/016b2410f1bc4753ba3339aa5caf3ffff5123dd7}{(Merabet et al., 2008-08)} by using TMS or other perturbation methods to test whether regions identified as containing 'sensory' representations actually causally contribute to hand-specific tactile processing. Sixth, expand the analysis beyond binary classification to examine the full representational geometry using techniques like multidimensional scaling or hierarchical clustering to better understand the relationship between sensory and perceptual representations. Finally, include proper control conditions such as non-letter tactile patterns or scrambled braille stimuli to ensure that observed effects are specific to meaningful braille processing rather than general tactile stimulation differences between hands.

[...More method-level reviews...]   
\end{tcolorbox}

\begin{tcolorbox}[enhanced, breakable, title = \system Example of a Dimension Contribution Evaluation]

[...More dimension-level reviews...]

\textbf{Dimension: Experimental Design}

\textbf{\textit{Strengths:}} The experimental design presents novelty through its systematic operationalization of sensory versus perceptual representations via hand-dependency. This approach is entirely absent from existing braille research, which typically focuses on stimulus categories \href{https://doi.org/10.1093/cercor/bhac216}{Tian et al., 2022}, orthographic contractions \href{https://doi.org/10.1101/2022.11.09.515790}{Liu et al., 2022}, or priming effects \href{https://doi.org/10.1162/jocn_a_01407}{[Rączy et al., 2019]}. The three-experiment structure combining fMRI spatial localization, EEG temporal dynamics, and behavioral similarity ratings provides comprehensive coverage of the research question that surpasses single-modality studies in the field. The specific focus on individual braille letters with controlled hand presentation using piezo-electric cells represents a novel experimental paradigm not found in existing literature, which typically examines words or uses traditional tactile presentation methods.

\textbf{\textit{Weaknesses:}} The experimental design, while novel in its specific implementation, addresses questions that overlap substantially with existing cross-modal plasticity research. Studies already examine how visual cortex processes tactile information in blind individuals (\href{https://doi.org/10.1093/cercor/bhy151}{[de Borst \& de Gelder, 2018]}; \href{https://doi.org/10.1101/2021.05.18.444648}{[Silson et al., 2021]}, and the basic framework of studying braille processing across different brain regions is well-established (\href{https://doi.org/10.1162/jocn_a_02041}{Liu et al., 2023}). The design's focus on letter-level processing, while providing experimental control, may limit ecological validity compared to word-level studies that better reflect natural braille reading behavior.

\textbf{\textit{Suggestions:}} The design could be strengthened by including a direct comparison condition with sighted individuals performing analogous visual letter discrimination tasks to better isolate the effects of visual deprivation. Adding a longitudinal component to track how hand-independence develops with braille reading experience could enhance the design's theoretical contribution. The authors should also consider including more complex braille stimuli (e.g., contracted braille) to bridge the gap between their controlled letter-level approach and real-world braille reading applications demonstrated in other studies (\href{https://doi.org/10.1101/2022.11.09.515790}{Liu et al., 2022}).

[...More dimension-level reviews...]
\end{tcolorbox}





\begin{center}
\small\textbf{List of AI papers in \benchmark.}
\end{center}
\label{tab:ai-paper-list}
\small
\begin{enumerate}\setlength\itemsep{2pt}\setlength\parskip{0pt}
\item \href{https://openreview.net/forum?id=YaRzuMaubS}{Marwa Abdulhai \textit{et al.}, ``Defining Deception in Decision Making''}\\
\item \href{https://openreview.net/forum?id=DlZ97cVwr0}{Philipp Guevorguian \textit{et al.}, ``Exploring the Recall of Language Models: Case Study on Molecules''}\\
\item \href{https://openreview.net/forum?id=DnBjhWLVU1}{Seungwon Oh \textit{et al.}, ``Recovering Plasticity of Neural Networks via Soft Weight Rescaling''}\\
\item \href{https://openreview.net/forum?id=qK6U4Ahfms}{Yuwei Yan \textit{et al.}, ``OpenCity: A Scalable Platform to Simulate Urban Activities with Massive LLM Agents''}\\
\item \href{https://openreview.net/forum?id=vVlNBaiLdN}{Moritz Glaser \textit{et al.}, ``ESMGain: Effective and Efficient Prediction of Mutation’s functional Effect via ESM2 Transfer Learning and robust Benchmarks''}\\
\item \href{https://openreview.net/forum?id=WxLwXyBJLw}{Svetlana Pavlova \textit{et al.}, ``Flow Matching for One-Step Sampling''}\\
\item \href{https://openreview.net/forum?id=NpsgBKlApa}{Sasan Tavakkol \textit{et al.}, ``Less is More: Adaptive Coverage for Synthetic Training Data''}\\
\item \href{https://openreview.net/forum?id=UoWslU6hsX}{Lorenzo Pacchiardi \textit{et al.}, ``100 instances is all you need: predicting LLM success by testing on a few instances''}\\
\item \href{https://openreview.net/forum?id=c8sEgxG2c0}{Zhihan Zhou \textit{et al.}, ``GenomeOcean: Efficient Foundation Model for Genome Generation''}\\
\item \href{https://openreview.net/forum?id=QiyQJqpcYe}{Eduardo Sánchez \textit{et al.}, ``Linguini: A benchmark for language-agnostic linguistic reasoning''}\\
\item \href{https://openreview.net/forum?id=DEOV74Idsg}{Zekun Li \textit{et al.}, ``MMSci: A Dataset for Graduate-Level Multi-Discipline Multimodal Scientific Understanding''}\\
\item \href{https://openreview.net/forum?id=hMEHnLJyrU}{Alexander Shypula \textit{et al.}, ``Does Instruction Tuning Reduce Diversity? A Case Study Using Code Generation''}\\
\item \href{https://openreview.net/forum?id=Ql7msQBqoF}{Naman Gupta \textit{et al.}, ``MAC-CAFE: Multi-actor, Centralized Critic Architecture for Feedback-driven Editing''}\\
\item \href{https://openreview.net/forum?id=hWF0HH8Rr9}{Magnus Müller \textit{et al.}, ``Large-Scale Multi-Agent Reinforcement Learning for Traffic Signal Optimization''}\\
\item \href{https://openreview.net/forum?id=gaa7gWPZBz}{Aohan Sun \textit{et al.}, ``Mitigating Privacy Risk of Adversarial Examples with Counterfactual Explanations''}\\
\item \href{https://openreview.net/forum?id=gtVo4xcpFI}{Gwok-Waa Wan \textit{et al.}, ``GenBen: A Genarative Benchmark for LLM-Aided Design''}\\
\item \href{https://openreview.net/forum?id=N0MnPLK6r7}{Kyeongrok Park \textit{et al.}, ``Toward Human-Interpretable Explanations in a Unified Framework for GNNs''}\\
\item \href{https://openreview.net/forum?id=o3V7OuPxu4}{Wenjie Tang \textit{et al.}, ``StarCraft II Arena: Evaluating LLMs in Strategic Planning, Real-Time Decision Making, and Adaptability''}\\
\item \href{https://openreview.net/forum?id=y15LAM4u0A}{Chen Gao \textit{et al.}, ``EmbodiedCity: A Benchmark Platform for Embodied Agent in Real-world City Environment''}\\
\item \href{https://openreview.net/forum?id=irCuIdCdAl}{Sungmin Han \textit{et al.}, ``Improving Transformer Interpretability with Activation Contrast-Based Attribution''}\\
\item \href{https://openreview.net/forum?id=9e5syenoVE}{Hong Xie \textit{et al.}, ``Multiple-play Stochastic Bandits with Prioritized Resource Sharing''}\\
\item \href{https://openreview.net/forum?id=skJLOae8ew}{Santiago Yeomans \textit{et al.}, ``From Abstract Noise to Architectural Form: Designing Diffusion Models for Efficient Floor Plan Generation''}\\
\item \href{https://openreview.net/forum?id=CFKZKjrQ5r}{Chinmay Mittal \textit{et al.}, ``FCoReBench: Can Large Language Models Solve Challenging First-Order Combinatorial Reasoning Problems?''}\\
\item \href{https://openreview.net/forum?id=Y9yQ9qmVrc}{Haihong Yang \textit{et al.}, ``scKGOT: Intercellular Signaling Inference with Knowledge Graph Optimal Transport for Single-cell Transcriptomics''}\\
\item \href{https://openreview.net/forum?id=gyTkfVYL45}{Changliang Zhou \textit{et al.}, ``ICAM: Rethinking Instance-Conditioned Adaptation in Neural Vehicle Routing Solver''}\\
\item \href{https://openreview.net/forum?id=7jDv1RrNQX}{Yisheng Xiao \textit{et al.}, ``Path Selection Makes BERT-family Good Generators''}\\
\item \href{https://openreview.net/forum?id=8sglLco8Ti}{Xiang Liu \textit{et al.}, ``ChunkKV: Semantic-Preserving KV Cache Compression for Efficient Long-Context LLM Inference''}\\
\item \href{https://openreview.net/forum?id=xcHIiZr3DT}{Teng Yan \textit{et al.}, ``Vision-Based Pseudo-Tactile Information Extraction and Localization for Dexterous Grasping''}\\
\item \href{https://openreview.net/forum?id=1S8ndwxMts}{Pavel Strashnov \textit{et al.}, ``Towards Robust Evaluation of Protein Generative Models: A Systematic Analysis of Metrics''}\\
\item \href{https://openreview.net/forum?id=glgvpS1dD1}{Zhenlei Wang \textit{et al.}, ``Robust Heterogeneous Treatment Effect Estimation under Covariate Perturbation''}\\
\item \href{https://openreview.net/forum?id=ctzGqxE3O0}{Yao Shiyi \textit{et al.}, ``BID: Broad Incremental for Android Malware Detection''}\\
\item \href{https://openreview.net/forum?id=to4PdiiILF}{Leo McKee-Reid \textit{et al.}, ``Honesty to Subterfuge: In-Context Reinforcement Learning Can Make Honest Models Reward Hack''}\\
\item \href{https://openreview.net/forum?id=v5BouOktUP}{Tingzhou Wei \textit{et al.}, ``Multivariate Time-series Forecasting with SPACE: Series Prediction Augmented by Causality Estimation''}\\
\item \href{https://openreview.net/forum?id=f6GMwpxXHG}{Anuradha Kumari \textit{et al.}, ``ZEPHYR GAN: REDEFINING GAN WITH FLEXIBLE GRADIENT CONTROL''}\\
\item \href{https://openreview.net/forum?id=WVzYMa68Of}{Andrei Chertkov \textit{et al.}, ``Tensor Train Decomposition for Adversarial Attacks on Computer Vision Models''}\\
\item \href{https://openreview.net/forum?id=S2WHlhvFGg}{Zhenghan Chen \textit{et al.}, ``Advancing Drug-Target Interaction Prediction via Graph Transformers and Residual Protein Embeddings''}\\
\item \href{https://openreview.net/forum?id=kvCKoKfqTd}{Zhenghan Chen \textit{et al.}, ``Non-Commutative Spectral Geometry for Adaptive Quantum-Classical Drug-Target Interaction Prediction''}\\
\item \href{https://openreview.net/forum?id=qeY25DwmKO}{Chris Cameron \textit{et al.}, ``Foundation Models for Boolean Logic''}\\
\item \href{https://openreview.net/forum?id=TLgDQ0Rr2Z}{Haoxuan Li \textit{et al.}, ``Principle Counterfactual Fairness''}\\
\item \href{https://openreview.net/forum?id=TkbjqexD8w}{Yuntian Wu \textit{et al.}, ``Invariant Spatiotemporal Representation Learning for Cross-patient Seizure Classification''}\\
\end{enumerate}

\begin{center}
\small\textbf{List of neuroscience papers in \benchmark.}
\end{center}
\label{tab:neuro-paper-list}
\small
\begin{enumerate}\setlength\itemsep{2pt}\setlength\parskip{0pt}\item \href{https://elifesciences.org/reviewed-preprints/89851v1#tab-content}{Benas \textit{et al.}, ``Modeled grid cells aligned by a flexible attractor''}\\
\item \href{https://elifesciences.org/reviewed-preprints/97793v1#tab-content}{Wittkamp \textit{et al.}, ``The neural dynamics of positive and negative expectations of pain''}\\
\item \href{https://elifesciences.org/reviewed-preprints/99862v1#tab-content}{Cui \textit{et al.}, ``Dysfunctional S1P/S1PR1 signaling in the dentate gyrus drives vulnerability of chronic pain-related memory impairment''}\\
\item \href{https://elifesciences.org/reviewed-preprints/92860v1#tab-content}{O’Leary \textit{et al.}, ``Natural forgetting reversibly modulates engram expression in hippocampal feedforward circuits''}\\
\item \href{https://elifesciences.org/reviewed-preprints/90930v1#tab-content}{Klaassen \textit{et al.}, ``Basolateral amygdala inhibition impairs updating of appetitive and aversive values by interacting with the prefrontal cortex''}\\
\item \href{https://elifesciences.org/reviewed-preprints/98148v1#tab-content}{Haupt \textit{et al.}, ``The transformation of sensory to perceptual braille letter representations in the visually deprived brain''}\\
\item \href{https://elifesciences.org/reviewed-preprints/97602v1#tab-content}{Liu \textit{et al.}, ``Cell class-specific long-range axonal projections of neurons in mouse whisker-related somatosensory cortices''}\\
\item \href{https://elifesciences.org/reviewed-preprints/106481v1#tab-content}{Campbell \textit{et al.}, ``Human single-neuron activity is modulated by intracranial theta burst stimulation of the basolateral amygdala''}\\
\item \href{https://elifesciences.org/reviewed-preprints/104543v1#tab-content}{Derkaloustian \textit{et al.}, ``Fine Touch Perception Relies on Frictional Instabilities''}\\
\item \href{https://elifesciences.org/reviewed-preprints/106614#tab-content}{Lee \textit{et al.}, ``The influence of temporal context on vision over multiple time scales''}\\
\item \href{https://elifesciences.org/reviewed-preprints/97326v1#tab-content}{Setogawa \textit{et al.}, ``Acquisition of auditory discrimination mediated by different processes through two distinct circuits linked to the lateral striatum''}\\
\item \href{https://elifesciences.org/reviewed-preprints/101105v1#tab-content}{Cooper \textit{et al.}, ``Ultraslow serotonin oscillations in the hippocampus delineate substates across NREM and waking''}\\
\item \href{https://elifesciences.org/reviewed-preprints/100947v1#tab-content}{Mollá–Albaladejo \textit{et al.}, ``Molecular characterization of gustatory second-order neurons reveals integrative mechanisms of gustatory and metabolic information''}\\
\item \href{https://elifesciences.org/reviewed-preprints/104222v1#tab-content}{Bloem \textit{et al.}, ``Dynamic estimation of the attentional field from visual cortical activity''}\\
\item \href{https://elifesciences.org/reviewed-preprints/107273#tab-content}{Xu \textit{et al.}, ``Neural Representation of Time across Complementary Reference Frames''}\\
\item \href{https://elifesciences.org/reviewed-preprints/98634v1#tab-content}{Rieser \textit{et al.}, ``Multifaceted Role of Galanin in Whole Brain Excitability''}\\
\item \href{https://elifesciences.org/reviewed-preprints/88376v1#tab-content}{Lu \textit{et al.}, ``The interplay between homeostatic synaptic scaling and homeostatic structural plasticity maintains the robust firing rate of neural networks''}\\
\item \href{https://elifesciences.org/reviewed-preprints/101850v1#tab-content}{Ecker \textit{et al.}, ``Assemblies, synapse clustering and network topology interact with plasticity to explain structure–function relationships of the cortical connectome''}\\
\item \href{https://elifesciences.org/reviewed-preprints/102475v1#tab-content}{Dash \textit{et al.}, ``Rules for reactivation across REM sleep microstates following sensory fear learning''}\\
\item \href{https://elifesciences.org/reviewed-preprints/103736v1#tab-content}{March \textit{et al.}, ``The Hungry Lens: Hunger Shifts Attention and Attribute Weighting in Dietary Choice''}\\
\item \href{https://elifesciences.org/reviewed-preprints/101959v1#tab-content}{Molkov \textit{et al.}, ``Introducing perturbations in point-process models of excitable systems''}\\
\item \href{https://elifesciences.org/reviewed-preprints/94835v1#tab-content}{Huang \textit{et al.}, ``Neural coding of multiple motion speeds in visual cortical area MT''}\\
\item \href{https://elifesciences.org/reviewed-preprints/107252#tab-content}{Liu \textit{et al.}, ``Striatal Crosstalk Between Dopamine and Serotonin Systems''}\\
\item \href{https://elifesciences.org/reviewed-preprints/104242v1#tab-content}{Kang \textit{et al.}, ``Rapid rebalancing of co-tuned ensemble activity in the auditory cortex''}\\
\item \href{https://elifesciences.org/reviewed-preprints/106387v1#tab-content}{Praegel \textit{et al.}, ``Age and Learning Shapes Sound Representations in Auditory Cortex During Adolescence''}\\
\item \href{https://elifesciences.org/reviewed-preprints/102756v1}{Zhang \textit{et al.}, ``Oxytocin restores context-specific hyperaltruistic preference''}\\
\item \href{https://elifesciences.org/reviewed-preprints/107370#tab-content}{Hall \textit{et al.}, ``A cortical–hippocampal communication undergoes rebalancing after new learning''}\\
\item \href{https://elifesciences.org/reviewed-preprints/107472#tab-content}{Zhang \textit{et al.}, ``Humans underestimate their body mass in microgravity''}\\
\item \href{https://elifesciences.org/reviewed-preprints/104008v1#tab-content}{Barnby \textit{et al.}, ``Self–other generalisation shapes social interaction and is disrupted in borderline personality disorder''}\\
\item \href{https://elifesciences.org/reviewed-preprints/100258v1#tab-content}{Tardiff \textit{et al.}, ``Normative evidence weighing and accumulation in correlated environments''}\\
\item \href{https://elifesciences.org/reviewed-preprints/106840#tab-content}{Wu \textit{et al.}, ``The Self-Interest of Adolescents Overrules Cooperation in Social Dilemmas''}\\
\item \href{https://elifesciences.org/reviewed-preprints/96718v1#tab-content}{Chen \textit{et al.}, ``Synchronous Ensembles of Hippocampal CA1 Pyramidal Neurons During Novel Exploration''}\\
\item \href{https://elifesciences.org/reviewed-preprints/105537v1#tab-content}{Wang \textit{et al.}, ``The relationship between cognitive abilities and mental health as represented by cognitive abilities at the neural and genetic levels of analysis''}\\
\end{enumerate}

\begin{center}
\small\textbf{List of biochemistry papers in \benchmark.}
\end{center}
\label{tab:biochem-paper-list}
\small
\begin{enumerate}\setlength\itemsep{2pt}\setlength\parskip{0pt}\item \href{https://elifesciences.org/reviewed-preprints/105225v1/reviews#tab-content}{Zhong \textit{et al.}, ``Modular DNA Barcoding of Nanobodies Enables Multiplexed in situ Protein Imaging and High-throughput Biomolecule Detection''}\\
\item \href{https://elifesciences.org/reviewed-preprints/96446v1#tab-content}{Majhi \textit{et al.}, ``Non-autonomous cell redox-pairs dictate niche homeostasis in multi-lineage stem populations''}\\
\item \href{https://elifesciences.org/reviewed-preprints/103432v1#tab-content}{Krwawicz \textit{et al.}, ``Introduction of cytosine-5 DNA methylation sensitizes cells to oxidative damage''}\\
\item \href{https://elifesciences.org/reviewed-preprints/105892v1#tab-content}{Xiu \textit{et al.}, ``Action mechanism of a novel agrichemical quinofumelin against \textit{Fusarium graminearum}''}\\
\item \href{https://elifesciences.org/reviewed-preprints/107123#tab-content}{Chang \textit{et al.}, ``Cancer cells differentially modulate mitochondrial respiration to alter redox state and enable biomass synthesis in nutrient-limited environments''}\\
\item \href{https://elifesciences.org/reviewed-preprints/106397#tab-content}{Mohanty \textit{et al.}, ``Deep Learning Reveals Endogenous Sterols as Allosteric Modulators of GPCRs''}\\
\item \href{https://elifesciences.org/reviewed-preprints/91046v1#tab-content}{Wang \textit{et al.}, ``Structure and evolution of Alanine/Serine Decarboxylases \& S-Adenosylmethionine Decarboxylases in plants''}\\
\item \href{https://elifesciences.org/reviewed-preprints/102422v1}{Wei \textit{et al.}, ``Crystal structure and catalytic mechanism of PL35 family glycosaminoglycan lyases with an ultrabroad substrate spectrum''}\\
\item \href{https://elifesciences.org/reviewed-preprints/106508#tab-content}{Govorunova \textit{et al.}, ``Blue-shifted ancyromonad channelrhodopsins for multiplex optogenetics''}\\
\item \href{https://elifesciences.org/reviewed-preprints/100914v1#tab-content}{He \textit{et al.}, ``Coordinated regulation of chemotaxis and resistance to copper by CsoR''}\\
\item \href{https://elifesciences.org/reviewed-preprints/104549#tab-content}{Jandu \textit{et al.}, ``Membrane mimetic thermal proteome profiling (MM-TPP) enables proteome-wide target engagement in membranes''}\\
\item \href{https://elifesciences.org/reviewed-preprints/97821v1#tab-content}{Chong \textit{et al.}, ``Establishing the foundations for a data-centric AI approach for virtual drug screening''}\\
\item \href{https://elifesciences.org/reviewed-preprints/103721#tab-content}{Schulze \textit{et al.}, ``Effects of residue substitutions on the cellular abundance of proteins
''}\\
\item \href{https://elifesciences.org/reviewed-preprints/102511#tab-content}{Maus \textit{et al.}, ``Screening the MMV Pathogen Box reveals the mitochondrial bc1-complex as a drug target in mature \textit{Toxoplasma gondii} bradyzoites''}\\
\item \href{https://elifesciences.org/reviewed-preprints/102680#tab-content}{Lefroncois \textit{et al.}, ``The Role of ATP Synthase Subunit e (ATP5I) in Mediating the Metabolic and Antiproliferative Effects of Biguanides''}\\
\item \href{https://elifesciences.org/reviewed-preprints/96841v1#tab-content}{Ntourmas \textit{et al.}, ``Endogenous oligomer formation underlies DVL2 condensates and promotes Wnt/$\beta$-catenin signaling''}\\
\item \href{https://elifesciences.org/reviewed-preprints/103699#tab-content}{Marks \textit{et al.}, ``Determining the off-target activity of antibiotics and novel translation initiation sites in mitochondria''}\\
\item \href{https://elifesciences.org/reviewed-preprints/102146#tab-content}{Leanza \textit{et al.}, ``Increased bone inflammation in type 2 diabetes and obesity correlates with Wnt signaling downregulation and reduced bone strength''}\\
\item \href{https://elifesciences.org/reviewed-preprints/101855#tab-content}{Zhang \textit{et al.}, ``Distinct mechanisms of inhibition of Kv2 potassium channels by tetraethylammonium and RY785''}\\
\item \href{https://elifesciences.org/reviewed-preprints/99809v1#tab-content}{Luo \textit{et al.}, ``Isobaric crosslinking mass spectrometry technology for studying conformational and structural changes in proteins and complexes''}\\
\item \href{https://elifesciences.org/reviewed-preprints/93673v1#tab-content}{Antenucci \textit{et al.}, ``Reassessing the substrate specificities of the major \textit{Staphylococcus aureus} peptidoglycan hydrolases lysostaphin and LytM''}\\
\item \href{https://elifesciences.org/reviewed-preprints/98885v1#tab-content}{Zhou \textit{et al.}, ``Structural insights into human propionyl-CoA carboxylase \ldots''}\\
\item \href{https://elifesciences.org/reviewed-preprints/99026v1#tab-content}{Liu \textit{et al.}, ``Genome-wide mapping of native co-localized G4s and R-loops \ldots''}\\
\item \href{https://elifesciences.org/reviewed-preprints/91168v1#tab-content}{D’Oliveira \textit{et al.}, ``Recognition and Cleavage of Human tRNA \ldots''}\\
\end{enumerate}

\begin{center}
\small\textbf{List of ecology papers in \benchmark.}
\end{center}
\label{tab:eco-paper-list}
\small
\begin{enumerate}\setlength\itemsep{2pt}\setlength\parskip{0pt}\item \href{https://elifesciences.org/reviewed-preprints/89485v1#tab-content}{Rucci \textit{et al.}, ``Effects of blood meal source and seasonality on reproductive traits of \textit{Culex quinquefasciatus} (Diptera: Culicidae)''}\\
\item \href{https://elifesciences.org/reviewed-preprints/105501#tab-content}{García-Ruiz \textit{et al.}, ``Fitness drivers of division of labor in vertebrates''}\\
\item \href{https://elifesciences.org/reviewed-preprints/95857v1#tab-content}{Nakagawa \textit{et al.}, ``An illusion of a macroecological law, abundance–occupancy relationship in birds''}\\
\item \href{https://elifesciences.org/reviewed-preprints/106655#tab-content}{Jiang \textit{et al.}, ``Assessing plant phenological changes based on drivers of spring phenology''}\\
\item \href{https://elifesciences.org/reviewed-preprints/105411v1#tab-content}{Howard–Spink \textit{et al.}, ``Old age variably impacts chimpanzee engagement and efficiency in stone tool use''}\\
\item \href{https://elifesciences.org/reviewed-preprints/107554#tab-content}{Rebindaine \textit{et al.}, ``Developmental constraints mediate the summer solstice reversal of climate effects on European beech bud set''}\\
\item \href{https://elifesciences.org/reviewed-preprints/107093v1#tab-content}{Smit \textit{et al.}, ``Risk-taking incentives predict aggression heuristics in female gorillas''}\\
\item \href{https://elifesciences.org/reviewed-preprints/104762v1#tab-content}{Croijmans \textit{et al.}, ``Strip cropping shows promising increases in ground beetle community diversity compared to monocultures''}\\
\item \href{https://elifesciences.org/reviewed-preprints/105585v1#tab-content}{Wang \textit{et al.}, ``Loss of olfaction reduces caterpillar performance and increases susceptibility to a natural enemy''}\\
\item \href{https://elifesciences.org/reviewed-preprints/98073v1#tab-content}{Tao \textit{et al.}, ``Partitioning changes in ecosystem productivity by effects of species interactions in biodiversity experiments''}\\
\item \href{https://elifesciences.org/reviewed-preprints/103339#tab-content}{Yang \textit{et al.}, ``Interpreting prediction intervals and distributions for biologically meaningful effects''}\\
\item \href{https://elifesciences.org/reviewed-preprints/91774v1#tab-content}{Gao \textit{et al.}, ``Pesticide-induced resurgence in brown planthoppers is mediated by action on a suite of genes that promote juvenile hormone biosynthesis \ldots''}\\
\item \href{https://elifesciences.org/reviewed-preprints/100041v1#tab-content}{Fargeot \textit{et al.}, ``Genetic diversity affects ecosystem functions across trophic levels \ldots''}\\
\item \href{https://elifesciences.org/reviewed-preprints/104700#tab-content}{Seltzer \textit{et al.}, ``Female Moths Incorporate Plant Acoustic Emissions into Their Oviposition Decision-Making Process''}\\
\item \href{https://elifesciences.org/reviewed-preprints/103406/reviews#tab-content}{Seguchi \textit{et al.}, ``Vasopressin 1a receptor antagonist disrupts male–male affiliative relationships formed by triadic cohabitation in large-billed crows''}\\
\item \href{https://elifesciences.org/reviewed-preprints/102991#tab-content}{Gatt \textit{et al.}, ``Integrating microscopy and transcriptomics from individual eukaryotic plankton (Ukiyo-e-Seq)''}\\
\item \href{https://elifesciences.org/reviewed-preprints/92227v1#tab-content}{Rydhmer \textit{et al.}, ``Automating an insect biodiversity metric using distributed optical sensors: an evaluation across Kansas, USA cropping systems''}\\
\item \href{https://elifesciences.org/reviewed-preprints/101906v1#tab-content}{Diaz–Colunga \textit{et al.}, ``Full factorial construction of synthetic microbial communities''}\\
\item \href{https://elifesciences.org/reviewed-preprints/97298v1#tab-content}{Zhang \textit{et al.}, ``Neuropeptide bursicon and its receptor mediate the transition in seasonal polyphenism of \textit{Cacopsylla chinensis}''}\\
\item \href{https://elifesciences.org/reviewed-preprints/103971v1#tab-content}{Zhang \textit{et al.}, ``Birds migrate longitudinally in response to the resultant Asian monsoons of the Qinghai–Tibet Plateau uplift''}\\
\end{enumerate}


\end{document}